%% file: root.tex
\documentclass[journal]{IEEEtran}
\usepackage[pdftex]{graphicx}
\usepackage{booktabs}
\usepackage{moreverb}
\usepackage{url}
\usepackage{amsmath}
\usepackage{bm}
\usepackage[caption=false]{subfig}
\usepackage{color}
\usepackage{amsfonts}
\usepackage{amssymb}
\usepackage{multicol}
\usepackage{mathtools}
\usepackage[usenames,dvipsnames]{xcolor}
\usepackage{algpseudocode}
\usepackage{algorithm}
\usepackage[colorlinks,bookmarksopen,bookmarksnumbered,citecolor=red,urlcolor=red]{hyperref}
\usepackage{cleveref}
\usepackage{siunitx}
\usepackage{glossaries}
	\newacronym{mpc}{MPC}{model prective control}
	\newacronym{oc}{OC}{optimal control}
	\newacronym{sfc}{SFC}{state-feedback controller}
	\newacronym{wbc}{WBC}{whole-body controller}
	\newacronym{ilqr}{iLQR}{iterative linear-quadratic regulator}
	\newacronym{admm}{ADMM}{alternating direction method of multipliers}
	\newacronym{ddp}{DDP}{differential dynamic programming}
	\newacronym{nlp}{NLP}{nonlinear programming}
	\newacronym{gn}{GN}{Gauss-Newton}
	\newacronym{qp}{QP}{quadratic programming}
	\newacronym{com}{CoM}{center of mass}
	\newacronym{zmp}{ZMP}{zero moment point}
	\newacronym{aba}{ABA}{articulated body algorithm}
	\newacronym{rnea}{RNEA}{recursive Newton-Euler algorithm}
	\newacronym{ccrbi}{CCRBI}{centroidal composite rigid body inertia}
	\newacronym{fk}{FK}{forward kinematics}
	\newacronym{cwc}{CWC}{contact wrench cone}
	\newacronym{ekf}{EKF}{extended Kalman filter}
	\newacronym{lf}{LF}{left-front}
	\newacronym{rf}{RF}{right-front}
	\newacronym{lh}{LH}{left-hind}
	\newacronym{rh}{RH}{right-hind}

\hypersetup
{
	pdftitle = {Agile Maneuvers in Legged Robots: a predictive control approach},
	pdfauthor = {Carlos Mastalli},
	pdfsubject = {T-RO paper},
	pdfkeywords = {agile maneuvers, model predictive control, state-feedback control, legged robots, full dynamics},
	pdftoolbar = true,
	colorlinks = true,
	linkcolor = black,
	citecolor = black,
	urlcolor = black,
}
\newcommand{\sref}[1]{Section~\ref{#1}}
\newcommand{\fref}[1]{Fig.~\ref{#1}}
\newcommand{\eref}[1]{Eq.~(\ref{#1})}

\newcommand{\aref}[1]{Appendix~\ref{#1}}

\newlength{\tempdima}
\newcommand{\rowname}[1]{\rotatebox{0}{\makebox[\tempdima][c]{(\footnotesize #1)}}}
\newcommand{\orcid}[1]{\href{https://orcid.org/#1}{\includegraphics[width=0.6em]{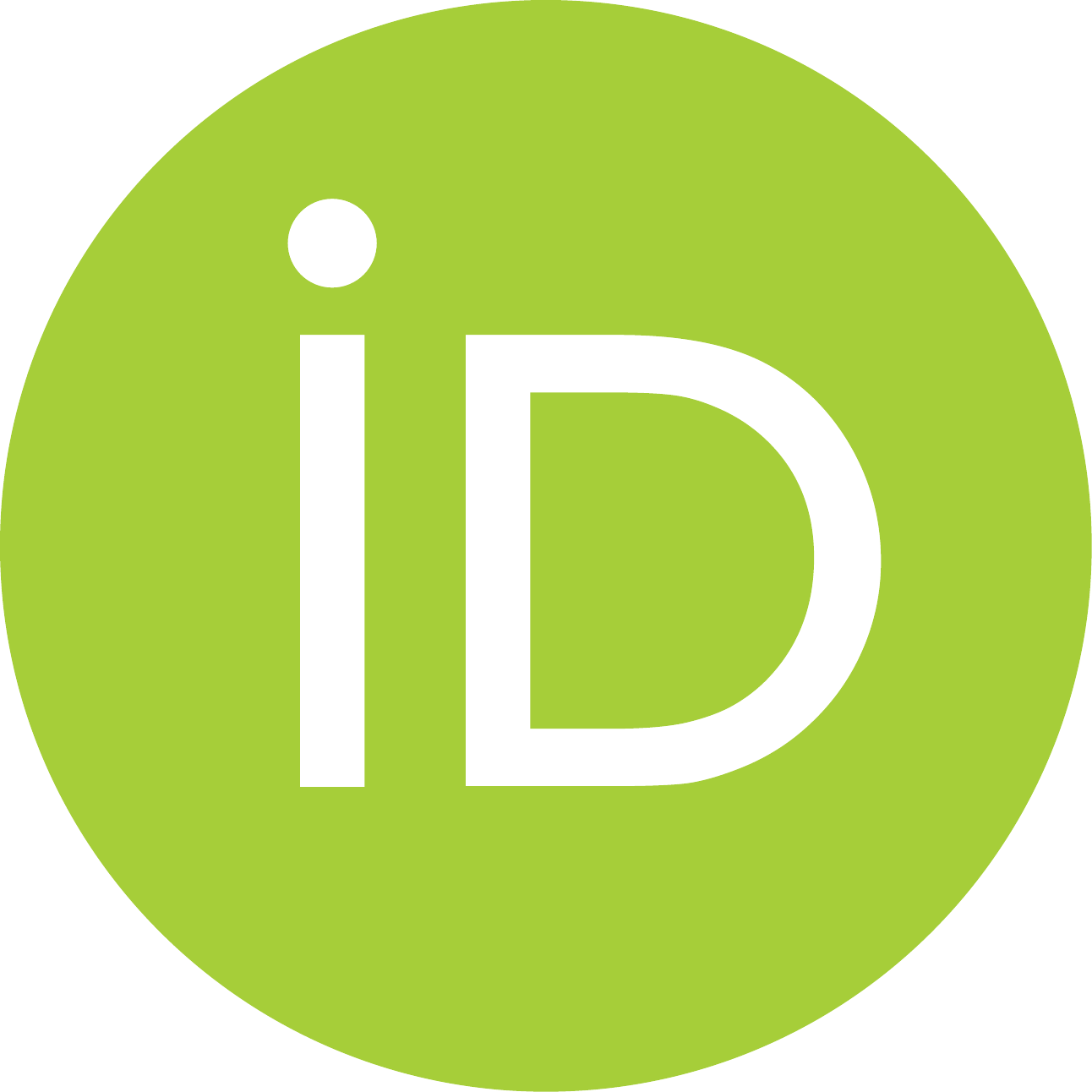}}}

\title{Agile Maneuvers in Legged Robots:\\ a Predictive Control Approach}
\author{
	Carlos Mastalli$^*$\orcid{0000-0002-0725-4279}\quad
	Wolfgang Merkt$^*$\orcid{0000-0003-3235-4906}\quad
	Guiyang Xin\orcid{0000-0003-0921-7487}\\
	Jaehyun Shim\orcid{0000-0002-8008-7120}\quad
	Michael Mistry\orcid{0000-0003-3979-922X}\quad
	Ioannis Havoutis\orcid{0000-0002-4371-4623}\quad
	Sethu Vijayakumar\orcid{0000-0003-0649-7241}
\thanks{%
This research was supported by (1) the European Commission under the Horizon 2020 project Memory of Motion (MEMMO, project ID: 780684), (2) the Engineering and Physical Sciences Research Council (EPSRC) UK RAI Hub for Offshore Robotics for Certification of Assets (ORCA, grant reference EP/R026173/1), and (3) the Alan Turing Institute.
$^*$The first two authors contributed equally to this work.
\textit{(Corresponding author: Carlos Mastalli.)}
}
\thanks{
Carlos Mastalli is with the School of Engineering and Physical Sciences, Heriot-Watt University, U.K. (e-mail: \href{mailto:c.mastalli@hw.ac.uk}{c.mastalli@hw.ac.uk}).
}
\thanks{
Jaehyun Shim, Michael Mistry and Sethu Vijayakumar are with the School of Informatics, University of Edinburgh, U.K. (e-mail: \href{mailto:jshim@ed.ac.uk}{jshim@ed.ac.uk};
\href{mailto:mmistry@ed.ac.uk}{mmistry@ed.ac.uk}; \href{mailto:sethu.vijayakumar@ed.ac.uk}{sethu.vijayakumar@ed.ac.uk}).
}
\thanks{
Wolfgang Merkt and Ioannis Havoutis are with the Oxford Robotics Institute, Department of Engineering Science, University of Oxford, U.K. (e-mail: \href{mailto:wolfgang@robots.ox.ac.uk}{wolfgang@robots.ox.ac.uk}; \href{mailto:ioannis@robots.ox.ac.uk}{ioannis@robots.ox.ac.uk}).
}
\thanks{
Guiyang Xin is with the School of Optoelectronic Engineering and Instrumentation Science, Dalian University of Technology, Dalian, China. (e-mail: \href{mailto:guiyang.xin@dlut.edu.cn}{guiyang.xin@dlut.edu.cn}).
}
}

\begin{document}

\maketitle

\input{src/0_abstract}

\input{src/1_introduction}

\input{src/2_pipeline}

\input{src/3_contact_phase_mpc}

\input{src/4_control}

\input{src/5_results}

\input{src/6_conclusion}

\bibliography{reference}

\appendices

\input{src/7_appendices}

\input{src/8_bios}

\end{document}

%% file: src/0_abstract.tex
\begin{abstract}%
    Planning and execution of agile locomotion maneuvers have been a longstanding challenge in legged robotics.
    It requires to derive motion plans and local feedback policies in real-time to handle the nonholonomy of the kinetic momenta.
    To achieve so, we propose a hybrid predictive controller that considers the robot’s actuation limits and full-body dynamics.
    It combines the feedback policies with tactile information to locally predict future actions.
    It converges within a few milliseconds thanks to a feasibility-driven approach.
    Our predictive controller enables ANYmal robots to generate agile maneuvers in realistic scenarios.
    A crucial element is to track the local feedback policies as, in contrast to whole-body control, they achieve the desired angular momentum.
    To the best of our knowledge, our predictive controller is the first to handle actuation limits, generate agile locomotion maneuvers, and execute optimal feedback policies for low level torque control without the use of a separate whole-body controller.
\end{abstract}
    

%% file: src/1_introduction.tex
\begin{figure}
  \centering
  \href{https://youtu.be/R8lFti7x5N8}{\includegraphics[width=0.99\columnwidth]{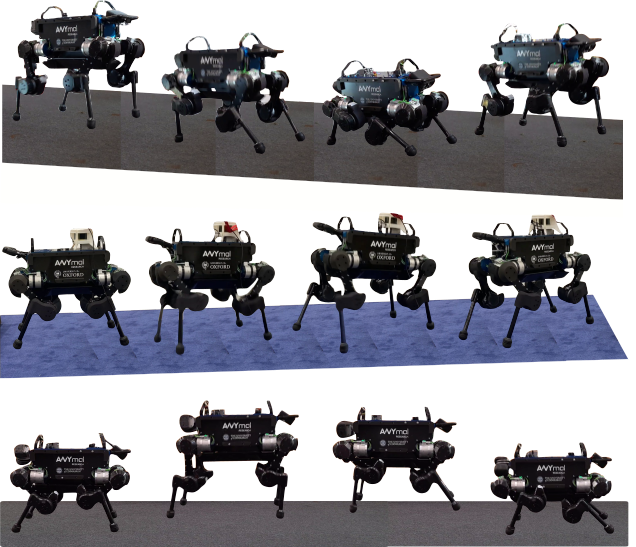}}
  \caption{Our hybrid predictive controller generating different agile maneuvers on the ANYmal B robot.
  In the top clip, ANYmal jumped diagonally twice.
  In the middle clip, ANYmal jumped four times with a rotation of 30 degrees each.
  In the bottom clip, ANYmal jumped 15cm forward.
  All the maneuvers were computed using our \textsc{Box-FDDP} solver, which efficiently handles torque limits and dynamic feasibility, and executed with our Riccati-based state-feedback controller.
  Each contact switching (landing phase) is described using impulse dynamics.
  To watch the video, click the figure or see \texttt{\url{https://youtu.be/R8lFti7x5N8}}}
  \label{fig:cover}
\end{figure}

\section{Introduction}
\IEEEPARstart{M}{odel} Predictive Control (\acrshort{mpc}) is a powerful tool to synthesize in real-time motions and controls through task goals (cost / optimality) and constraints (system dynamics, interaction constraints, etc.)~\cite{mayne-automatica00}.
One of the benefits of state-feedback predictive control is to handle nonholonomic systems, as according to Brockett theorem~\cite{brockett-dgct83} we cannot stabilize these systems with continuous time-invariant feedback control laws.
In rigid body systems, nonholonomics constraints appear in the conservation of the angular momentum~\cite{wieber-fmbr05}.
This means that, during flying motions, the robot cannot properly regulate the angular momentum with instantaneous control actions (e.g.,~\cite{wensing-icra13,herzog-iros14,delprete-tro16,shamelmastalli-ral19}).
As a consequence, it is important to consider the full-body dynamics for the generation, and not only execution, of agile maneuvers such as jumps and flying trots.
Furthermore, \acrshort{mpc} approaches are recognized to react against unforeseen disturbances and model errors in a wide range of situations.
To give an example in the case of legged locomotion, predictive control helps to compensate strong push forces~\cite{wieber-ichr06} or to produce safer motions by simply minimizing the jerk of the \gls{com}~\cite{wieber-iros08}.
It promises to leverage motion plans and controls together while increasing the overall robustness and exploiting the robot's capabilities for \textit{agile maneuvers}.
However, as briefly justified below, we can potentially enable those promises by 1) employing the full-body dynamics of the robot, by 2) following the optimal policy, and by 3) improving the globalization of algorithms for numerical optimal control.
We propose a framework that addresses these challenges in this work.
\fref{fig:cover} shows the generation of different agile maneuvers that are only possible when our \acrshort{mpc} (i) considers the torque limits and (ii) tracks the optimal policy via our Riccati-based state-feedback predictive controller.

\subsection{Brief justification}
To date, most of the locomotion frameworks follow a \textit{motion-planning} paradigm (e.g.,~\cite{kuindersma-ar16,mastalli-tro20}), in which an instantaneous controller tracks motion plans computed from reduced-order dynamics (e.g.,~\cite{bledt-icra20,farshidian-ichr17,winkler-ral18}).
In this control paradigm, the strategy is to encode, in the instantaneous controller, a hand-tuned control policy (i.e., a set of tasks and weights) that depends on motion plans often computed assuming reduced-order dynamics.
However, when considering plans computed using the full-body dynamics, this leads to a sub-optimal policy as the instantaneous controller \textit{re-writes} the motion plans provided by the full-dynamics motion planner.
This sub-optimality arises as the hand-tuned policies do not align with the optimal policy, especially if there is \textit{nonholonomy in the dynamics}~\cite{wieber-ichr06} as it describes a non-conservative vector field (i.e., path-dependent field).
This highlights the importance of following the optimal policy based on the full-body dynamics, which constitutes what we call as \textit{motion-control} paradigm.
In this work, we propose this new control paradigm and show that increases \textit{tracking performance} while maintaining the robot's compliance.

Additionally, reduced-order models neglect the nonholonomy in the dynamics, the robot's actuation and kinematic limits by approximating the kinetic angular momenta (i.e., the Euler equation of motion).
This assumption is very restrictive as it requires actuators with high bandwidth and torque limits such as the ones developed for the Atlas robot.
It limits the application of these \textit{simplified} predictive controllers to robots with light limbs.
As we wish to generate agile maneuvers in any kind of robot, it is important to develop predictive control algorithms that consider the robot's full dynamics, its actuation limits, and compute locally optimal policies.
In this work, we also propose a \textit{full-dynamics~\acrshort{mpc}} that can generate agile maneuvers efficiently.
To the best of our knowledge, our predictive controller is the first to handle nonholonomy in the kinetic momenta, torque limits and friction-cone constraints, which allows the ANYmal robot to perform multiple jumps.

\subsection{Related work}
The nonholonomy in the kinetic momenta, introduced above, is related with the non-Euclidean nature of rigid body systems~\cite{marsden-mcg98}.
This effect can be handled (at least) using the centroidal dynamics together with the robot's full kinematics~\cite{orin-ar13}, as the momentum of a rigid body system depends on the robot's configuration and velocity.
However, most state-of-the-art approaches in trajectory optimization neglect the momentum components associated with the robot's limbs, as it leads to a simpler and smaller optimization problem for motion generation~\cite{carpentier-icra16,ponton-ichr16,farshidian-ichr17} and contact planning~\cite{aceituno_cabezas-ral18,winkler-ral18,wang-iros20}.
In contrast, the earliest effort that adopts the centroidal dynamics for trajectory optimization~\cite{dai-ichr14} has a high computational cost, and it cannot be used in predictive control.
Its high computation cost is due to (i) formulating the contact dynamics using complementarity constraints (also reported by others~\cite{yunt-amc06,posa-ijrr14,mastalli-icra16}), and (ii) building upon a general-purpose nonlinear optimization software (which cannot exploit the problem structure efficiently).
In a follow-up work, Herzog et al.~\cite{herzog-iros16} proposed a kinodynamic optimal control formulation that significantly reduces the computational cost by employing \gls{admm} and hybrid dynamics.
The former aspect reduces the computation time needed by off-the-shelf solvers, while the latter one avoids the use of complementarity constraints.
But, even though that this approach decreases the computation time, it seems to be still high for predictive control applications as the results were reported in the context of trajectory optimization only.
This is also the case for recent works that similarly split the nonconvex optimal control problem into two convex and alternating sub-problems~\cite{ponton-tro21,shah-iros21}.
It is also important to note that, with the exception of~\cite{ponton-tro21}, these approaches do not consider actuation limits.

Many works have recently demonstrated dynamic locomotion realized using \acrshort{mpc} with reduced-order models such as inverted pendulum model~\cite{wieber-ichr06,wieber-iros08} or the single-rigid body dynamics \cite{dicarlo-iros18,bledt-icra20,rathod-access21} with full kinematics~\cite{farshidian-ichr17,grandia-iros19}.
With former models based on inverted pendulum, we can improve the robot's stability compared with the \gls{zmp} approach~\cite{kajita-icra03}, as we could incorporate further constraints.
With latter approaches based on single-rigid body dynamics, legged robots can achieve more dynamic motions through fast resolution of \textit{direct transcription} problems.\footnote{For more details about direct transcription formulation see~\cite{betts-bookoptctrl}.}
However, these simplifications ``hold'' for legged robots where (i) the mass and inertia of the limbs are negligible compared to the torso (e.g., MIT's Mini Cheetah) and (ii) the actuation capabilities are unlimited with infinite torque limits and bandwidth (e.g., Boston Dynamics's Atlas).
Conversely, some authors have resorted to model the leg inertia explicitly through lumped mass or learned models \cite{ahn-frai-21,wang-arxiv-21}.
Again, neither of these models can explicitly include actuation constraints/capabilities in the formulation.
Indeed, this is possible with the \textit{full rigid body dynamics}, or its projection to the~\gls{com} point that it is well-known as centroidal dynamics~\cite{orin-ar13}.

Predictive control with full rigid body dynamics is challenging, as it is a large nonlinear optimization problem that needs to be solved within a few milliseconds~\cite{betts-bookoptctrl}.
Nevertheless, Tassa et al.~\cite{tassa-iros12} demonstrated in simulation that it is feasible to deploy \acrshort{mpc} based on the robot's full dynamics with the \gls{ilqr}\footnote{The \gls{ilqr} is also commonly recognized to be a \gls{ddp} algorithm~\cite{mayne-66} with Gauss-Newton approximation.} algorithm~\cite{li-icinco04}.
Note that this approach employed MuJoCo's smooth and invertible contact model~\cite{todorov-icra14}.
Later, the first experimental validation was performed on the HRP-2 humanoid robot, but for kinematic manipulation tasks only~\cite{koenemann-iros15} as MuJoCo's contact dynamics seems to be not accurate enough for locomotion.
In the same vein, a similar approach was developed for quadrupedal locomotion~\cite{neunert-ral18}, in which a linear spring-damper system models the contact interactions.
This approach enables the \acrshort{mpc} controller to automatically make and break contact, albeit with the complexity of tuning the hyperparameters of the contact model that artificially creates non-zero phantom forces during non-contact phases.
These phantom forces predicts unrealistic motions, especially during flying phases, and it is unlikely to generate agile and complex locomotion maneuvers under this assumption.
Indeed, the authors have demonstrated their approach in simple motions: trotting and walking in flat terrain.
Again, both approaches neglect the robot's actuation limits as they rely on the~\gls{ilqr} algorithm.

The promising efficiency of \gls{ilqr}-like solvers has attracted special attention in the legged robotics community.
These techniques can be classified as \textit{indirect methods}, since they are a numerical implementation of the Pontryagin's minimun principle~\cite{betts-bookoptctrl}.
Their efficiency is due to the induced sparse structure via the ``Bellman's principle of optimality''.
Concretely speaking, it breaks the large matrix factorization, needed often in nonlinear programming solvers such as \textsc{Snopt}~\cite{gill-siam05}, \textsc{Knitro}~\cite{byrd-knitro06}, and \textsc{Ipopt}~\cite{wachter-mp06}, into a sequence of much smaller matrix factorizations.
These features make it appealing to real-time \acrshort{mpc}, in particular due to the fact that libraries for linear algebra, such as \textsc{Eigen}~\cite{eigenweb}, are often very efficient to handle small and dense matrix factorizations.\footnote{With dense matrices, we can further leverage vectorization and code generation of matrix operations (e.g.,~\cite{carpentier-19sii}), which improves efficiency on modern CPUs.}
However, their main limitations are 1) their inability to handle arbitrary constraints, and 2) the poor globalization of indirect approaches.
Note that the \textit{algorithm's globalization}, or limited capability to converge from arbitrary starting points, is of great interest, as it prohibits the optimization of agile maneuvers such as jumps~\cite{mastalli-icra20}.

Most recent works focus on the former limitations, which derive from basic notions proposed in~\cite{tassa-icra14}.
Indeed, this work showed the connection with the control Hamiltonian minimization, typically found in textbooks~(e.g.,~\cite{betts-bookoptctrl}), which can be extended for handling arbitrary constraints as proposed in~\cite{xie-icra17}.
Unfortunately, this extension increases the computation time by up to four times.
To improve this, Howell et al.~\cite{howell-19} proposed an augmented Lagrangian approach to boost the computational efficiency, which was later extended in~\cite{kazdadi-icra-21,sleiman-icra-21}.
On the other hand, for improving the poor globalization, a modification of the Riccati equations to account for \textit{lifted dynamics} was proposed in~\cite{giftthaler-iros18}.
Note that the term lifted is a name coined by~\cite{albersmeyer-siam10}, and it refers to \textit{gaps} or \textit{defects} produced between multiple-shooting nodes.
Then, Mastalli et al.~\cite{mastalli-icra20} included a modification of the forward pass that numerically matches the gap contraction expected by a \textit{direct multiple-shooting} method with only equality constraints.
This approach drives the iteration towards the \textit{feasibility} of the problem.
It can be seen as a \textit{direct-indirect} hybridization, and it can handle control limits~\cite{mastalli-underreview20}.

\subsection{Contribution}
The scope of this paper is to achieve agile maneuvers in legged robots.
For that, we first start by justifying --with a fresh view-- the importance of full-body dynamics and predictive control.
We then propose a hybrid model predictive control approach that performs such kind of motions by considering the robot's full dynamics.
To enhance convergence needed in agile maneuvers, we formulate and solve a \textit{feasibility driven problem}.
Our predictive controller computes motion plans and feedback policies given a predefined sequence of contact phases in real-time.
It uses tactile information to locally predict future actions and controls.
We introduce the friction/wrench cone constraints based on our hybrid paradigm~\cite{budhiraja-ichr18,mastalli-icra20}.
To achieve high control frequency rates while considering the robot's full dynamics, our approach employs multiple-shooting to enable parallelization in the computation of analytical derivatives.

Secondly, we reduce the reliance on separate tracking controllers and exploit the local feedback control laws synthesized at the motion planning level.
As mentioned above, this constitutes a \textit{motion-control paradigm}, in which in contrast to~\cite{grandia-iros19} we follow both: optimal trajectories and policies.
To the best of our knowledge, we demonstrate the first application of local feedback policies (Riccati gains) for low-level torque control in hardware experiments for locomotion tasks.
The key benefit hereto is that the optimal feedback gains increase the tracking performance and compliance as they follow the optimal and non-conservative vector field defined by the nonholonomic of the robot's dynamics.
Our experimental analysis compares our Riccati-based state-feedback control approach against a hierarchical whole-body controller based on the robot's inverse dynamics.
We validate our pipeline in a series of maneuvers of progressively increasing complexity on the ANYmal B and C quadrupeds.\footnote{B and C refer to the version of the ANYmal robots, for example both are shown in \fref{fig:highly_dyn_maneuvers}.}

%% file: src/2_pipeline.tex
\section{Overview}\label{sec:overview}
In this section, we first justify the importance of using full-body dynamics and predictive control, which is particularly important for the generation of agile maneuvers (\sref{sec:role_full_dynamics}). 
Later, we briefly describe our locomotion pipeline (\sref{sec:pipeline}).

\subsection{The role of full-body dynamics}\label{sec:role_full_dynamics}
It is often assumed that centroidal optimization on the robot's trunk is the key ingredient for dynamic and versatile locomotion, as the centroidal momentum of the robot’s trunk dominates the centroidal dynamics in robots with light-weight legs and arms.
However, if we translate this assumption in nature, then we would not see cats recover from a fall or astronauts that reorient themselves in zero-gravity.
Technically speaking, this assumption removes the nonholonomic nature of the centroidal kinetic momentum of rotation of rigid body systems~\cite{wieber-fmbr05}, which in the case of flight phases is expressed as (top row):
\begin{equation}\label{eq:am_conservation}
\sum_{i=0}^{N_b}
\overbrace{\begin{bmatrix}[\mathbf{x}_i-\mathbf{p}_G]_\times m_i & \mathbf{R}_i\mathbf{\bar{I}}^{cm}_i \\
m_i & 0
\end{bmatrix}}^{{}^i\mathbf{X}_G^\top\mathbf{I}_i}
\begin{bmatrix}
\mathbf{\dot{x}}_i \\ \boldsymbol{\omega}_i
\end{bmatrix} = \text{constant},
\end{equation}
where $\mathbf{x}_i$ describes the position of the body $i$ with mass $m_i$, $\mathbf{\bar{I}}^{cm}_i$ corresponds to its rotational inertia matrix about body's \gls{com}, $\mathbf{R}_i$ defines its orientation, $\mathbf{p}_G$ is the~\gls{com} position of the rigid body system, $[\cdot]_\times$ constructs a skew-symmetric matrix, and $N_b$ is the number of rigid bodies.
Note that the first term in the LHS term can be computed using spatial algebra\footnote{For more details about spatial algebra see~\cite{featherstone-rbdbook}.} as ${}^i\mathbf{X}_G^\top\mathbf{I}_i$, where 
${}^i\mathbf{X}_G$ is the spatial transform between the total~\gls{com} and the body $i$, and $\mathbf{I}_i$ is its spatial inertial.
Additionally, for sake of simplicity, we describe the bodies' configuration and velocity using maximal coordinates expressed in their~\gls{com} points.

\begin{figure}[tbp]
    \centering\begin{tabular}{cc}
    \rowname{a} & \href{https://youtu.be/R8lFti7x5N8?t=10}{\raisebox{-.5\height}{\includegraphics[width=0.9\columnwidth]{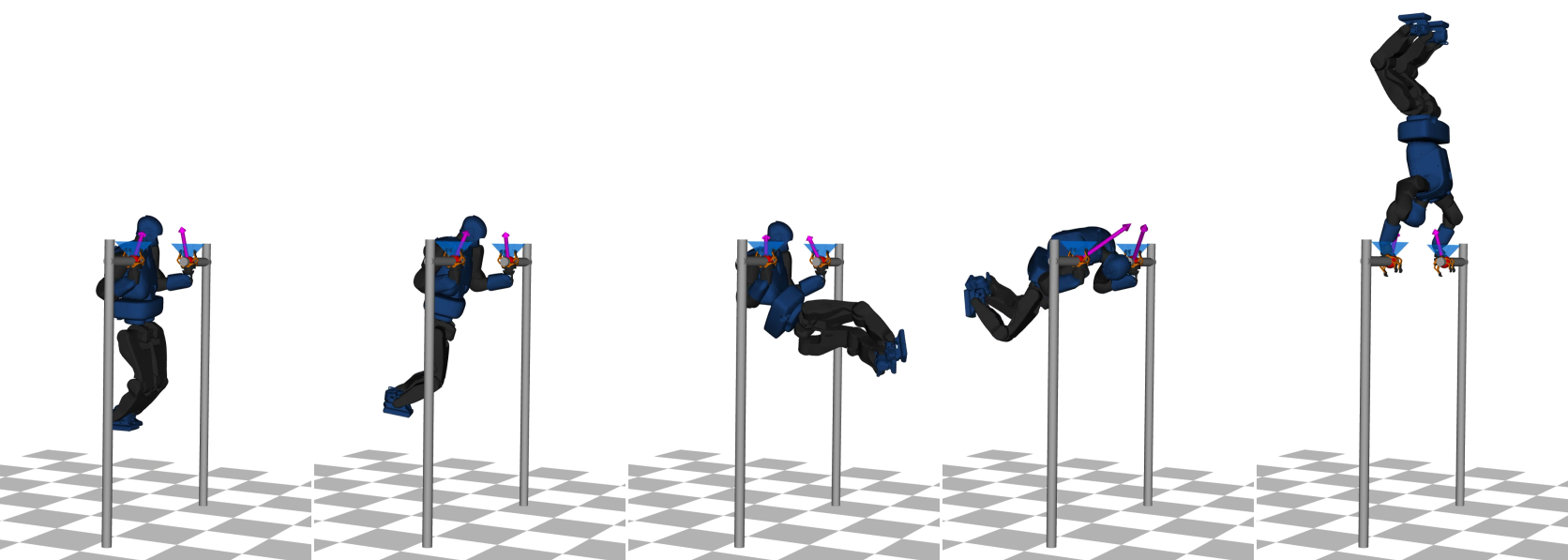}}}\\
    \rowname{b} & \href{https://youtu.be/R8lFti7x5N8?t=10}{\raisebox{-.5\height}{\includegraphics[width=0.93\columnwidth]{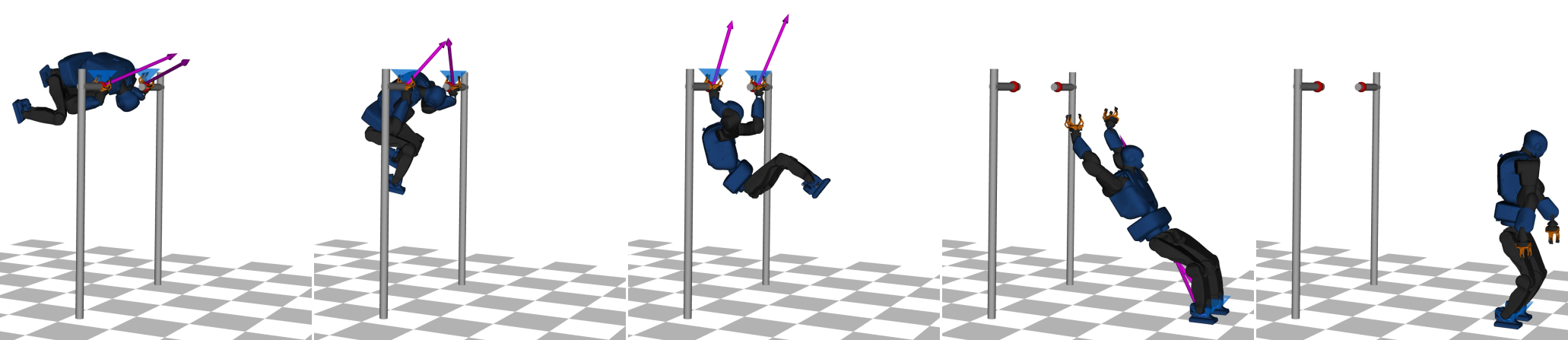}}}
    \end{tabular}
    \caption{Building-up momentum and modifying the locked inertia tensor from first principles of optimization.
    Both sequence of motions are optimized using the full dynamics and joint limits of the Talos robot.
    To reduce the torques required, the Talos robot performs:
    (a) two swinging motions needed to achieve the handstand posture.
    (b) reduction of the locked inertia tensor needed to achieve a quicker jumping velocity and desired landing place.}
    \label{fig:humanoid_handstand}
\end{figure}

\subsubsection{Nonholonomic effect in flight phases}
As the kinetic momentum of rotation has to be constant, we observe that it is possible to modify the robot's posture by changing its \textit{locked inertial tensor} (name coined in~\cite{marsden-mcg98}).
This statement becomes clear if we express~\eref{eq:am_conservation} in terms of the \textit{average spatial velocity} introduced in~\cite{orin-ar13}, i.e.,
\begin{equation*}
\overbrace{\left[\sum_{i=0}^{N_b}
{}^i\mathbf{X}_G^\top\mathbf{I}_i{}^i\mathbf{X}_G\right]}^{\mathbf{I}_G}
\mathbf{v}_G = \text{constant},
\end{equation*}
where $\mathbf{v}_G$ is the average spatial velocity, and $\mathbf{I}_G$ is the (spatial) locked inertial tensor -- also known as the~\gls{ccrbi} matrix~\cite{orin-ar13}.
Moreover, the top row of differential constraint in~\eref{eq:am_conservation} tells us that, to reach a desired orientation of the robot, each body needs to follow a specific path (i.e., path-dependent field) as it defines a \textit{nonholonomic evolution}.
It also tells us that to reach a desired footstep location, from a flight phase, we need to plan the entire kinematic path of each body along the (constant) angular momentum generated in the take-off phase.
Indeed, \fref{fig:humanoid_handstand}b shows that if we optimize the motion with the full-body dynamics (using our formulation in~\sref{sec:oc_formulation}), then we can generate a maneuver that increases the jumping velocity by reducing the locked inertia tensor.
This increment in the jumping velocity is crucial for landing in the desired location (\SI{1.8}{\meter} from the bars).

\subsubsection{Nonholonomic effect in contact phases}
The limitations of the above-mentioned assumption go beyond maneuvers with flight phases.
The nonholonomic structure of the kinetic momentum of rotation in~\eref{eq:am_conservation} remains similar when a robot is walking, if we note its connection with the Euler equation:
\begin{equation}\label{eq:euler_equation}
\frac{d}{dt}\overbrace{\left[\sum_{i=0}^{N_b}(\mathbf{A}_{\mathbf{k}_G})_i\mathbf{v}_i\right]}^{\mathbf{k}_G} = \sum_{j=0}^{N_f} [\mathbf{p}_j - \mathbf{p}_G]_\times\boldsymbol{\lambda}_j + \boldsymbol{\tau}_j,
\end{equation}
with
\begin{equation*}
(\mathbf{A}_G)_i\coloneqq
\begin{bmatrix}
(\mathbf{A}_{\mathbf{k}_G})_i \\ (\mathbf{A}_{\mathbf{l}_G})_i
\end{bmatrix} =
{}^i\mathbf{X}_G^\top\mathbf{I}_i
\mathbf{J}_i
\end{equation*}
as the row of the centroidal momentum matrix~\cite{orin-ar13} associated to the joint $i$, $\mathbf{J}_i$ as the body Jacobian, $\mathbf{v}_i$ relates the velocity of body $i$ to its predecessor,
$\mathbf{p}_j$ as the position of the contact $j$, $\boldsymbol{\lambda}_j$ and $\boldsymbol{\tau}_j$ as the forces and torques acting on this contact, with $N_f$ number of contacts.
Note that we use minimal coordinates in~\eref{eq:euler_equation}, as it is the conventional description used in the \textit{centroidal dynamics}~\cite{orin-ar13}.
Additionally, the term within the square brackets in the LHS of~\eref{eq:euler_equation} is the centroidal angular momentum $\mathbf{k}_G$.

\begin{figure}[bp]\centering
    \includegraphics[width=0.9\columnwidth]{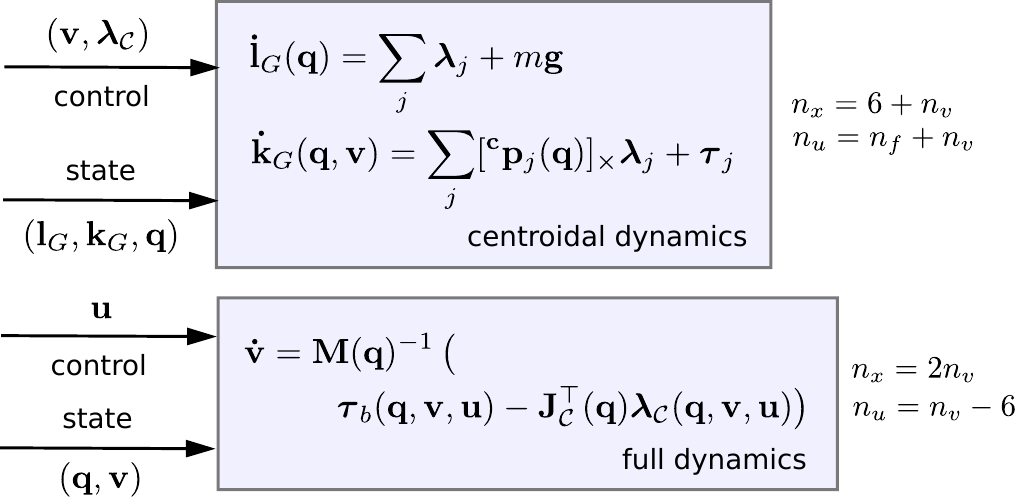}
    \caption{System dimension of centroidal and full-body dynamics.
    The complexity of algorithms for resolving optimal control problems depends on the system dimension.
    In the centroidal dynamics, the state of the system is described by the linear $\mathbf{l}_G\in\mathbb{R}^3$, angular $\mathbf{k}_G\in\mathbb{R}^3$, and generalized position $\mathbf{q}\in\mathbb{R}^{n_v}$, and its inputs are the generalized velocity $\mathbf{v}\in\mathbb{R}^{n_v}$ and contact forces $\boldsymbol{\lambda}_\mathcal{C}\in\mathbb{R}^{n_f}$.
    Instead in the full-body dynamics, the state is defined by the generalized position and velocity, and its input are joint torque commands $\mathbf{u}\in\mathbb{R}^{n_v-6}$ (assuming full actuation), where $\boldsymbol{\tau}_b$ defines the force-bias vector as defined in~\eref{eq:force_bias}.
    It means that if we use the centroidal dynamics, then the complexity of optimal control problem scales with the number of contact forces.
    Consequentially, its dimension (i.e., $n_x+n_u$) is lower compared with the full-body dynamics when $n_f\leq n_v - 12$.
    In case of quadruped robots, the dimension of the centroidal dynamics is equals to or greater than full-body dynamics if two feet or more are in contact (i.e., in trotting, pacing, and walking gaits).}
    \label{fig:dimension_full_vs_centroidal_dyn}
\end{figure}

From the structure of the Euler equation, we observe that, given a well-designed kinematic path, a rigid body system can build up angular momentum to reduce the required contact forces and torques.
Reduction on the contact forces and torques translates into lower joint torques, which increases the robots' capability to perform agile and complex maneuvers.
This is also true for robots with light-weight legs, as we will report in~\sref{sec:results} that our ANYmal robot could not execute multiple jumps if we do not track the angular momentum accurately.
\fref{fig:humanoid_handstand}a shows how the Talos humanoid robot builds up momentum by swinging motions twice.
This strategy allows the robot to reach the handstand posture.

\subsubsection{System actuation considerations}
As seen before, we can account for the nonholonomic effect using the centroidal dynamics too (not only using the \textit{full-body dynamics}).
Indeed, the centroidal model projects the rigid body dynamics to its~\gls{com} point given the full kinematics of the system~\cite{orin-ar13}.
Its inputs are the contact forces, instead of the joint torques as in the full-body dynamics.
It means that the dimension of both models remains similar,
with the dimension of the \textit{centroidal dynamics} potentially being smaller or larger depending on whether the dimension of the contact forces is lower or larger than the joint commands (see~\fref{fig:dimension_full_vs_centroidal_dyn}).
Therefore, using centroidal dynamics together with full kinematics does not provide a significant reduction on the computation time needed to solve optimal control problems, as the complexity of these algorithms are often related with the dimension of the system and horizon, and their nonlinearities remain similar.
For instance, the algorithm complexity of a Riccati recursion is approximately $\mathcal{O}(N(n_x + n_u)^3)$~\cite{frison-cca13}, where $N$ is the optimization horizon, and $n_x + n_u$ is the dimension of the dynamics computed from its state and control dimensions.

Retrieving the joint torques from the centroidal dynamics requires to re-compute the \textit{manipulator dynamics}, which adds an unnecessary extra computation.
Additionally, it defines highly nonlinear torque-limits constraints, which are not yet possible to handle (by state-of-the-art solvers, e.g.,~\cite{howell-19}) fast enough for~\acrshort{mpc}.
A workaround is to compute the joint torques $\mathbf{u}$ by assuming a quasi-static condition, e.g., $\mathbf{u}=\mathbf{J}_\mathcal{C}^\top\boldsymbol{\lambda}_\mathcal{C}$.
However, these assumptions do not hold in highly-dynamic maneuvers.
Therefore, it is more convenient to develop a predictive controller based on the full-body dynamics rather than based on the centroidal dynamics.
Below, we present a locomotion pipeline in which our predictive controller (core element) produces local policies using the robot's full dynamics.
As the joint torques are explicitly represented, we can directly handle actuation limits.

\subsection{Locomotion pipeline}\label{sec:pipeline}
\begin{figure}[tbp]
    \includegraphics[width=\columnwidth]{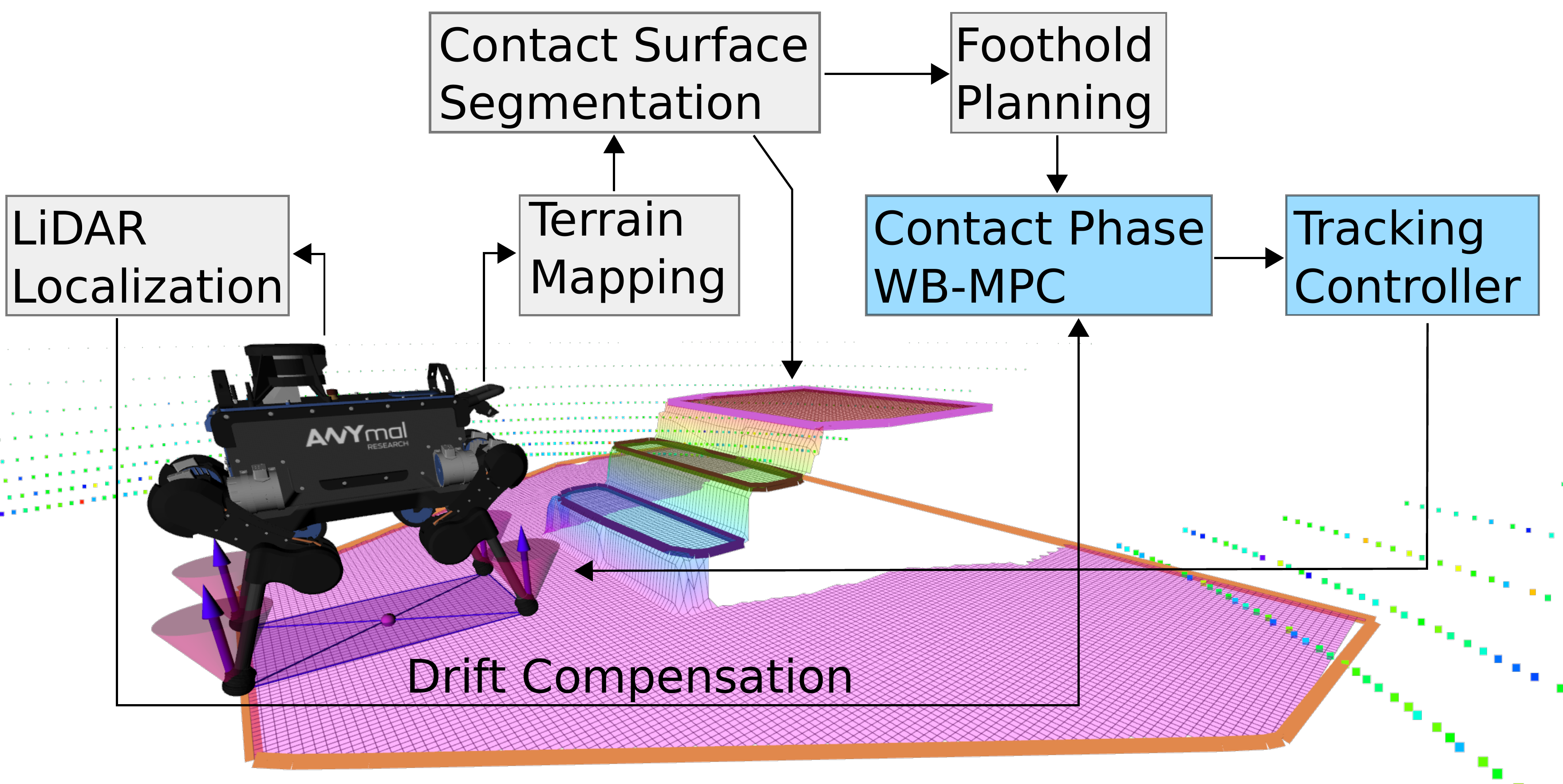}
    \caption{Overview of the locomotion pipeline. 
    The hybrid whole-body \acrshort{mpc} computes new policies (\SIrange{50}{100}{\hertz}) and transmits it to a tracking controller which closes the loop with state estimation and sensing (\SI{400}{\hertz}).
    We use LiDAR localization to compensate for state estimation drift in the \acrshort{mpc} (\sref{sec:state_estimation}).
    We integrate open-source modules for terrain mapping (\texttt{elevation\_mapping}), contact surface segmentation (\texttt{plane\_seg}), and foothold planning (\textrm{SL1M}~\cite{tonneau-icra20}) to automatically synthesize contact schedules based on navigation goals (\sref{sec:perceptive_footstep_planning}).
    Our key contributions, highlighted in blue, are described in~\sref{sec:contact_phase_mpc} and \ref{sec:sfc}. %
    }
    \label{fig:system_architecture}
\end{figure}

The heart of our locomotion pipeline is a predictive controller that computes local policies using the robot's full dynamics.
We formulate it as a hybrid optimal control problem, which includes contact and impulse dynamics defined from a contact schedule (\sref{sec:contact_phase_mpc}).
Then, a Riccati-based state-feedback controller follows the local and optimal policies computed by the predictive controller (\sref{sec:control_policy}).
Our full-dynamics~\acrshort{mpc} is designed to track footstep plans given a navigation goal.
\fref{fig:system_architecture} depicts the different components of our locomotion pipeline, which we briefly describe below.

\subsubsection{Hybrid MPC}
The hybrid~\acrshort{mpc}, described in the next section, re-computes the motion plans and policies at a fixed update frequency.
It accounts for delays in the communication, as it predicts the initial state of system at the time at which the tracking controller receives the updated policy.
This predicted state considers drift corrections from LiDAR scans (localization) as well.
In most of our experiments, our~\acrshort{mpc} operates at \SI{50}{\hertz}, despite that it is capable to run at \SI{100}{\hertz} on laptop hardware with \SI{1.25}{\sec} of optimization horizon.
We discretize this horizon at \SI{100}{\hertz}, which represents $125$ nodes.
The update rates of our \textit{full-dynamics~\acrshort{mpc}} are thus comparable to update rates reported in literature using single-rigid body dynamics with full kinematics (e.g.,~\cite{farshidian-ichr17}), which experimentally supports the thesis exposed in the previous sub-section.

\subsubsection{Tracking controller} \label{sec:tracking_controller}
Our predictive controller computes local feedback policies at \SI{50}{\hertz} (or up to \SI{100}{\hertz}) and sends these -- i.e., reference state and contact force trajectories along with time-varying feedback gain matrices -- to the tracking controller for a given control horizon.
The control horizon is typically around \SI{40}{\milli\second} (or 4 nodes), as it is enough to cope with unexpected delays in the communication.
Our \acrshort{mpc} and tracking controller communicate via \textsc{ROS-TCP/IP}.
The tracking controller runs at \SI{400}{\hertz}, and upsamples the reference state trajectory based on a system rollout/simulation.
Our Riccati-based state-feedback control constitutes what we refer as \textit{motion-control system}, which we describe in~\sref{sec:sfc}. Crucially, the feedback gain matrices are used directly to modify joint state and torque references for the low-level joint controllers without the requirement of a separate tracking controller as in~\cite{grandia-iros19}.
However, for comparison purpose, we also develop an instantaneous controller based on inverse dynamics and hierarchical quadratic programming (\sref{sec:wbc}).

\subsubsection{Perceptive footstep planning}\label{sec:perceptive_footstep_planning}
As a first step, we reconstruct a local terrain (elevation) map~\cite{fankhauser-clawar14} using depth sensors and then segment various convex patches.
As a second step, we compute the optimal foothold locations and contact patches from the segmented terrain using SL1M~\cite{tonneau-icra20,song-ral21} and a contact-surface segmentation.
Both footholds and convex patches are optimized based on a quasi-static assumption of the dynamics and chosen to achieve a desired navigation goal and passed to the contact-phase~\acrshort{mpc}.
With it, we compute the contact schedule needed for the predictive controller.
Below, we describe our full-dynamics~\acrshort{mpc}.

%% file: src/3_contact_phase_mpc.tex
\section{Hybrid MPC} \label{sec:contact_phase_mpc}
In this section, we first introduce the hybrid dynamics with time-based contact gains used in our model predictive controller (\sref{sec:hybrid_dynamics}).
Then, we describe our \textit{contac-phase} optimal control formulation focused on \textit{feasibility} (\sref{sec:oc_formulation}).

\subsection{Hybrid dynamics with time-based contact gain events}\label{sec:hybrid_dynamics}
We consider the hybrid dynamics of a rigid contact interaction as:
\begin{equation}
    \label{eq:hybrid_contact_dynamics}
    \begin{aligned}
        \mathbf{\dot{v}} &= \mathbf{f}_p(\mathbf{q},\mathbf{v},\mathbf{u}), \,\,\, p\in\mathcal{P}=\{1,2,\cdots,N_p\}\\
        \mathbf{v}^+ &= \boldsymbol{\Delta}_{p}(\mathbf{q},\mathbf{v}^-),
    \end{aligned}
\end{equation}
where $\mathbf{f}_p(\cdot)$ describes the evolution of the rigid body system under a set of predefined rigid contacts in phase $p$,
$\boldsymbol{\Delta}_{p}(\cdot)$ defines a rigid contact-gain transition between $p+1$ and $p$ phases, $N_p$ defines a number of contact phases, $\mathbf{q}$ and $\mathbf{v}$ are the generalized position and velocity, $\mathbf{v}^+$ and $\mathbf{v}^-$ represent the generalized velocities post- and pre-contact gain respectively, and $\mathbf{u}$ is the joint torque command.
Further details about the rigid body dynamics and its contact-gain transition models will be formally described in the next section.
Note that the term \textit{contact-gain} is a name coined in~\cite{featherstone-rbdbook}.

\begin{figure}[t]
    \centering
    \includegraphics[width=1.\columnwidth]{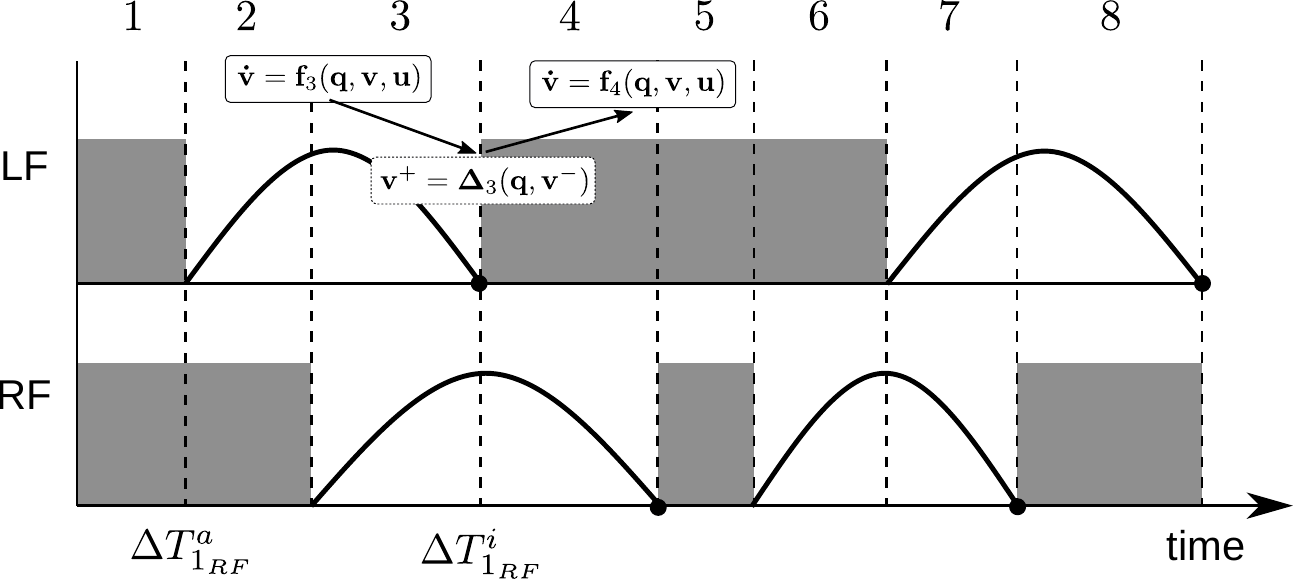}
    \caption{The hybrid dynamics and its time-based contact gain events given a known contact schedule.
    The diagram describes a dynamic walking gait for a bipedal robot, where LF and RF are the left and right feet.
    The active phases are marked with the gray block; while during the inactive phase a desired swing motion is defined.
    The black dot indicates when an inactive phase gains a contact, e.g., $\mathbf{v^+}=\boldsymbol{\Delta}_3(\mathbf{q},\mathbf{v^-})$.
    The contact gain instance creates a transition between different models, e.g., between $\mathbf{\dot{v}}=\mathbf{f}_3(\mathbf{q},\mathbf{v},\mathbf{u})$ and $\mathbf{\dot{v}}=\mathbf{f}_4(\mathbf{q},\mathbf{v},\mathbf{u})$.
    Note that these transitions do not depend on the control inputs.
    To put it clearly and succinctly, we only describe the duration of the first active and inactive phases for each contact/foot.
    Finally, given this desired contact schedule, the objective of our predictive controller is to find a sequence of whole-body reference motions, control commands and feedback gains.}
    \label{fig:hybrid_dynamics}
\end{figure}

For each contact, there are two main phases: active and inactive.
The active phase restricts the motion of its frame of reference, while the inactive one allows it to move freely towards a desired location.
The contact-gain transitions occur at each
\begin{equation*}
    \sum_{j=1}^{N_c} \Delta T^a_j + \Delta T^i_j\,:\, j\in\{1,2,\cdots,N_c\},
\end{equation*}
where $T^a_j$ and $T^i_j$ are the time duration of the active and inactive phases for the motion segment $j$, and $N_c$ defines the number of contacts (i.e., feet).
\fref{fig:hybrid_dynamics} depicts the hybrid dynamics given a predefined contact schedule: timings and references for the swing-foot motions.
With it, the objective of our predictive controller is to find a sequence of reference motions, control commands and feedback gains within a receding horizon of length duration $N$.
This receding horizon could be shorter than the contact schedule duration.

\subsubsection{Contact schedule}
\begin{figure}[t]
    \centering
    \includegraphics[width=1.\columnwidth]{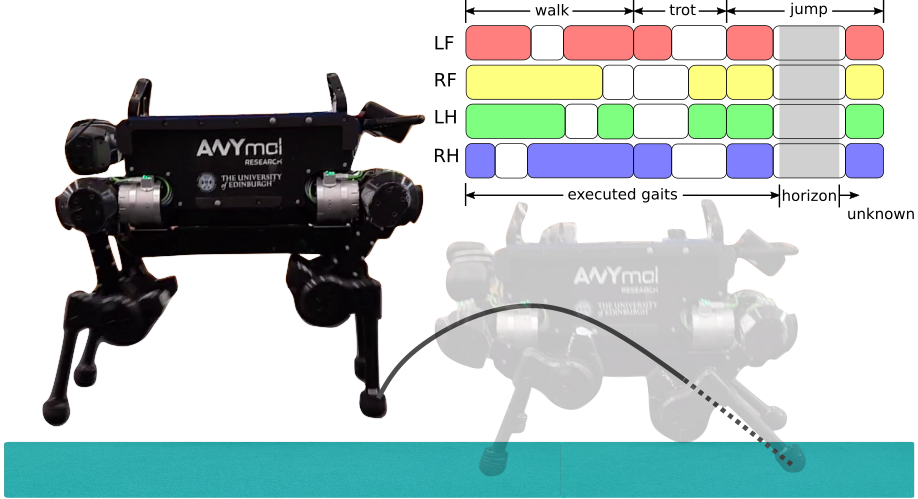}
    \caption{Contact schedule, gait duration, and optimization horizon for a sequence of walking, trotting and jumping gaits.
    The color boxes in the gait diagram (top-right) describe the activation of each foot of the ANYmal robot: red, yellow, green and blue are for the \gls{lf}, \gls{rf}, \gls{lh}, and \gls{rh} feet, respectively.
    The ANYmal robot is flying while planning in real-time its pre-landing motion (partially displayed by the continuous gray curve) along the current optimization horizon depicted by the gray box, after it walked and trotted.
    The gait diagram in the top-right shows the current optimization horizon (gray box), executed motion and unconsidered gaits.}
    \label{fig:contact_schedule}
\end{figure}

The contact schedule defines a set of active and inactive phases for each contact independently.
It allows us to define any kind of locomotion gait, or generally speaking, any kind of motion through contacts.
We define a number of motion segments $N_c$ for each contact $c$, with each segment including an active phase first, followed by an inactive phase.
Similarly to~\cite{winkler-ral18}, an active or inactive phase could be removed if we define its timings $\Delta T^a_j=0$ or $\Delta T^i_j=0$, respectively.
This particular motion description can be used to define any kind of behavior as the reference swing-foot motions allow us to use an optimization horizon that is different from the locomotion gait duration (\fref{fig:contact_schedule}).

\subsection{Hybrid \gls{oc} formulation}\label{sec:oc_formulation}
Given a predefined contact sequence and timings, at each \acrshort{mpc} step we formulate a hybrid optimal control problem that exploits the sparsity of the contact dynamics and its derivatives\footnote{We exploit the sparse structure of $\mathbf{M}$ and $\mathbf{J}_c$ during the matrix-matrix multiplications needed for the factorization.} as:
\begin{equation}\label{eq:oc_problem}
\begin{aligned}
\min_{\mathbf{x}_s,\mathbf{u}_s}
&\hspace{-2.em}
& & \hspace{-0.75em}\sum_{k=0}^{N-1} \|\mathbf{q}_k\ominus\mathbf{q}^{ref}\|^2_\mathbf{Q}+\|\mathbf{v}_k\|^2_\mathbf{N}+\|\mathbf{u}_k\|^2_\mathbf{R}+\|\boldsymbol{\lambda}_{\mathcal{C}_k}\|^2_\mathbf{K} \hspace{-8.em}&\\
& \hspace{-1.5em}\textrm{s.t.} & &\hspace{-1em}\text{if $k$ is a contact-gain transition:}\\
& & & \mathbf{q}_{k+1} = \mathbf{q}_{k},\\
& & & \left[\begin{matrix}\mathbf{v}_{k+1} \\ -\boldsymbol{\Lambda}_{\mathcal{C}_k}\end{matrix}\right] =
\left[\begin{matrix}\mathbf{M}_k & \mathbf{J}^{\top}_{\mathcal{C}_k} \\ {\mathbf{J}_{\mathcal{C}_k}} & \mathbf{0} \end{matrix}\right]^{-1}
\left[\begin{matrix}\boldsymbol{\tau}^\mathcal{I}_{b_k} \\ -\mathbf{a}^\mathcal{I}_{\mathcal{C}_k} \\\end{matrix}\right], \hspace{-1em}&\textrm{(impulse dyn.)}\\
& & & \hspace{-1em}\textrm{else:}\\
& & & \mathbf{q}_{k+1} = \mathbf{q}_k \oplus \int_{t_k}^{t_k+\Delta t_k}\hspace{-2em}\mathbf{v}_{k+1}\,dt, &\\
& & & \mathbf{v}_{k+1} = \mathbf{v}_k + \int_{t_k}^{t_k+\Delta t_k}\hspace{-2em}\mathbf{\dot{v}}_k\,dt, &\textrm{(integrator)}\\
& & & \left[\begin{matrix}\mathbf{\dot{v}}_k \\ -\boldsymbol{\lambda}_{\mathcal{C}_k}\end{matrix}\right] =
\left[\begin{matrix}\mathbf{M}_k & \mathbf{J}^{\top}_{\mathcal{C}_k} \\ {\mathbf{J}_{\mathcal{C}_k}} & \mathbf{0} \end{matrix}\right]^{-1}
\left[\begin{matrix}\boldsymbol{\tau}^\mathcal{C}_{b_k} \\ -\mathbf{a}^\mathcal{C}_{\mathcal{C}_k} \\\end{matrix}\right], \hspace{-1em}&\textrm{(contact dyn.)}\\
& & & \hspace{-1em}\mathbf{C}\boldsymbol{\lambda}_{\mathcal{C}_k} \geq \mathbf{c}, &\textrm{(friction-cone)}\\
& & & \hspace{-1em}\log{({}^\mathcal{W}\mathbf{p}_{\mathcal{G}_k}^{-1}\cdot {}^\mathcal{W}\mathbf{p}^{ref}_{{\mathcal{G}_k}})} = \mathbf{0}, &\textrm{(contact pos.)}\\
& & & \hspace{-1em}{}^\mathcal{W}\mathbf{\dot{p}}_{\mathcal{G}_k}^{-1}- {}^\mathcal{W}\mathbf{\dot{p}}^{ref}_{{\mathcal{G}_k}} = \mathbf{0}, &\textrm{(contact velocity)}\\
& & & \hspace{-1em}\mathbf{\underline{x}} \leq \mathbf{x}_k \leq \mathbf{\bar{x}}, &\textrm{(state bounds)}\\
& & & \hspace{-1em}\mathbf{\underline{u}} \leq \mathbf{u}_k \leq \mathbf{\bar{u}}, &\textrm{(control bounds)}
\end{aligned}
\end{equation}
where the state $\mathbf{x}=(\mathbf{q},\mathbf{v})\in X$ lies in a differential manifold (with dimension $n_x$) formed by the configuration point $\mathbf{q}$ and its tangent vector $\mathbf{v}\in\mathbb{R}^{n_v}$, the control $\mathbf{u}\in\mathbb{R}^{n_u}$ are the input joint torque commands, $\mathcal{C}_k$ and $\mathcal{G}_k$ define respectively the set of active and swing contacts given the running node $k$, $\mathbf{M}\in\mathbb{R}^{n_{v}\times n_{v}}$ is the joint-space inertia matrix, $\mathbf{J}_\mathcal{C}\in\mathbb{R}^{n_f\times n_v}$ is the contact Jacobian (full-rank matrix), $\boldsymbol{\lambda}_\mathcal{C},\boldsymbol{\Lambda}_\mathcal{C}\in\mathbb{R}^{n_f}$ is the contact force / impulse, $\mathbf{a}^\mathcal{C,I}_\mathcal{C}\in\mathbb{R}^{n_f}$ is the desired acceleration in the contact / impulse constraint space, $\boldsymbol{\tau}^\mathcal{C,I}_b\in\mathbb{R}^{n_v}$ is the force-bias vector which its definition changes for contact and impulse dynamics, ${}^\mathcal{W}\mathbf{p}_{\mathcal{G}}$, ${}^\mathcal{W}\mathbf{\dot{p}}_{\mathcal{G}}$ are the contact position and velocity, and $(\mathbf{C},\mathbf{c})$ describe the linearized friction cone or wrench cone~\cite{caron-icra15}.
Note that $n_v$, $n_u$ and $n_f$ describe the dimension of the rigid body system, number of actuated joints and contact forces, respectively.
Furthermore, the configuration point lies in a differential manifold defined as $\mathbf{q}\in\mathbb{SE}(3)\times\mathbb{R}^{n_j}$, where $n_j$ is the number of articulated joints, and its integration and difference operations are defined by $\oplus$ and $\ominus$.

\subsubsection{Feasibility-driven formulation}
Our optimal control formulation focuses on finding feasible solutions as we define regularization (cost) terms only.
Indeed, as written in~\eref{eq:oc_problem}, we regularize the system configuration, velocity, control and contact forces around their origin, with the exception of the configuration where we use the reference robot's posture $\mathbf{q}^{ref}\in\mathbb{SE}(3)\times\mathbb{R}^{n_j}$.
We define a set of diagonal weight matrices for the posture regularization $\mathbf{Q}\in\mathbb{R}^{n_v\times n_v}$, velocity regularization $\mathbf{N}\in\mathbb{R}^{n_v\times n_v}$, torque regularization $\mathbf{R}\in\mathbb{R}^{n_u\times n_u}$, and contact-force regularization $\mathbf{K}\in\mathbb{R}^{n_f\times n_f}$.
Our formulation can also regularize the controls around the quasi-static torques: \begin{equation*}\|\mathbf{S}\mathbf{u}_k-\mathbf{g}(\mathbf{q}_k) + \mathbf{J}_{\mathcal{C}_k}^\top\boldsymbol{\lambda}_{\mathcal{C}_k}\|^2_\mathbf{N},
\end{equation*}
where $\mathbf{S}\in\mathbb{R}^{n_u\times n_v}$ is the selection matrix, $\mathbf{g}(\cdot)\in\mathbb{R}^{n_v}$ is the gravity vector.
We efficiently compute the partial derivatives of the gravity vector based on the analytical derivatives of the~\gls{rnea} algorithm~\cite{carpentier-rss18}.

\subsubsection{Efficient hybrid dynamics}\label{sec:derivatives_of_hybrid_dynamics}
As explained in~\cite{udwadia-92}, the evolution of the systems dynamics can be described by (i) the stack of contact Jacobian expressed in the local frame $\mathbf{J}_\mathcal{C}$ and defined by the $\mathcal{C}$ indices, (ii) the desired acceleration in the constraint space $\mathbf{a}_\mathcal{C}$ (with Baumgarte stabilization~\cite{baumgarte-72}), (iii) the joint-space inertia matrix $\mathbf{M}$, and (iv) the force-bias vector $\boldsymbol{\tau}_b$.
For \textit{regular} phases the force-bias term and the constrained acceleration are defined as
\begin{eqnarray}\nonumber\label{eq:force_bias}
\boldsymbol{\tau}^\mathcal{C}_b\coloneqq\mathbf{S}\mathbf{u} + \mathbf{h(q,v)},\,\,\,\,
\mathbf{a}^\mathcal{C}_\mathcal{C}\coloneqq\mathbf{\dot{J}}_\mathcal{C}\mathbf{v} + \boldsymbol{\psi}(\mathbf{q},\mathbf{v}),
\end{eqnarray}
where $\mathbf{h}\in\mathbb{R}^{n_v}$ describes the Coriolis and gravitational forces, and $\boldsymbol{\psi}$ improves the numerical integration through a Baumgarte stabilization strategy~\cite{baumgarte-72}.
Instead for \textit{contact-gain} events, the force-bias term and constrained acceleration describes an impulse instance, i.e.,
\begin{equation}
\left[\begin{matrix}\mathbf{v}^+ \\ -\boldsymbol{\Lambda}_\mathcal{C} \end{matrix}\right] =
\left[\begin{matrix}\mathbf{M} & \mathbf{J}^{\top}_\mathcal{C} \\ {\mathbf{J}_\mathcal{C}} & \mathbf{0} \end{matrix}\right]^{-1}
\left[\begin{matrix}\boldsymbol{\tau}^\mathcal{I}_b \\ -\mathbf{a}^\mathcal{I}_\mathcal{C} \\\end{matrix}\right],
\label{eq:kkt_impulse}
\end{equation}
with
\begin{eqnarray}\nonumber\label{eq:force_bias_impulse}
\boldsymbol{\tau}^\mathcal{I}_b\coloneqq\mathbf{M}\mathbf{v}^-,\,\,\, \mathbf{a}^\mathcal{I}_\mathcal{C}\coloneqq e\mathbf{J}_\mathcal{C}\mathbf{v}^-,
\end{eqnarray}
where $\boldsymbol{\Lambda}_\mathcal{C}$ is the contact impulse and, $\mathbf{v}^-$ and $\mathbf{v}^+$ are the discontinuous changes in the generalized velocity (i.e., velocity before and after impact, respectively), and $e\in[0,1]$ is the restitution coefficient that considers compression / expansion.
Note that the impulse dynamics computes the post impact velocity $\mathbf{v}^+$, which we numerically integrate for obtaining the next configuration $\mathbf{q}$.

To increase efficiency, we do not factorize the entire matrix defined in the contact dynamics.
Instead, the system acceleration and contact forces can be expressed as:
\begin{eqnarray}\label{eq:impulse_dynamics}
    \mathbf{\dot{v}} &=& \mathbf{M}^{-1}\left(\boldsymbol{\tau}_b + \mathbf{J}^\top_\mathcal{C}\boldsymbol{\lambda}_\mathcal{C}\right),\nonumber\\
    \boldsymbol{\lambda}_\mathcal{C} &=& \widehat{\mathbf{M}}^{-1}(\mathbf{a}_\mathcal{C} - \mathbf{J}_\mathcal{C}\mathbf{M}^{-1}\boldsymbol{\tau}_b),
\end{eqnarray}
thanks to a blockwise matrix inversion, which is also commonly described in the numerical optimization literature as Schur complement factorization~\cite{nocedal-optbook}.
The method boils down to two Cholesky decompositions that compute $\mathbf{M}^{-1}$ and $\widehat{\mathbf{M}}^{-1}=\mathbf{J}_\mathcal{C}\mathbf{M}^{-1}\mathbf{J}_\mathcal{C}^\top$, where the latter term is known as the Schur complement.
$\widehat{\mathbf{M}}$ is also well-known in robotics as the operational space inertia matrix~\cite{khatib-jra87}.
We perform a similar procedure to compute the post-impact velocity and impulse in~\eref{eq:impulse_dynamics}.

\subsubsection{Analytical derivatives of hybrid dynamics}\label{sec:contact_derivatives}
We apply the chain rule to analytically compute the derivatives of the system acceleration and contact forces (or post-impact velocity and contact impulses).
As we express contact forces as accelerations in the contact constraint sub-space, we can indeed compute the derivatives of the contact-forces analytically.
These derivatives are needed to describe the friction-cone constraints, force and quasi-static torque regularization.
If we apply the chain rule, the first-order Taylor approximation of the system acceleration and contact forces are defined as:
\begin{eqnarray}
    \delta\mathbf{\dot{v}} =
    \overbrace{\left[\begin{matrix}\frac{\partial\mathbf{\dot{v}}}{\partial\boldsymbol{\tau}} & \frac{\partial\mathbf{\dot{v}}}{\partial\mathbf{a}_\mathcal{C}}\end{matrix}\right]
    \left[\begin{matrix}\frac{\partial\boldsymbol{\tau}}{\partial\mathbf{x}} \\ \frac{\partial\mathbf{a}_\mathcal{C}}{\partial\mathbf{x}}\end{matrix}\right]}^{\mathbf{f}_{\mathbf{x}}}\delta\mathbf{x}
    +
    \overbrace{\left[\begin{matrix}\frac{\partial\mathbf{\dot{v}}}{\partial\boldsymbol{\tau}} & \frac{\partial\mathbf{\dot{v}}}{\partial\mathbf{a}_\mathcal{C}}\end{matrix}\right]
    \left[\begin{matrix}\frac{\partial\boldsymbol{\tau}}{\partial\mathbf{u}} \\ \frac{\partial\mathbf{a}_\mathcal{C}}{\partial\mathbf{u}}\end{matrix}\right]}^{\mathbf{f}_{\mathbf{u}}}\delta\mathbf{u},\\
    \delta\boldsymbol{\lambda} =
    \overbrace{\left[\begin{matrix}\frac{\partial\boldsymbol{\lambda}_\mathcal{C}}{\partial\boldsymbol{\tau}} & \frac{\partial\boldsymbol{\lambda}_\mathcal{C}}{\partial\mathbf{a}_\mathcal{C}}\end{matrix}\right]
    \left[\begin{matrix}\frac{\partial\boldsymbol{\tau}}{\partial\mathbf{x}} \\ \frac{\partial\mathbf{a}_\mathcal{C}}{\partial\mathbf{x}}\end{matrix}\right]}^{\boldsymbol{\lambda}_{\mathbf{x}}}\delta\mathbf{x}
    +
    \overbrace{\left[\begin{matrix}\frac{\partial\boldsymbol{\lambda}_\mathcal{C}}{\partial\boldsymbol{\tau}} & \frac{\partial\boldsymbol{\lambda}_\mathcal{C}}{\partial\mathbf{a}_\mathcal{C}}\end{matrix}\right]
    \left[\begin{matrix}\frac{\partial\boldsymbol{\tau}}{\partial\mathbf{u}} \\ \frac{\partial\mathbf{a}_\mathcal{C}}{\partial\mathbf{u}}\end{matrix}\right]}^{\boldsymbol{\lambda}_{\mathbf{u}}}\delta\mathbf{u},
    \label{eq:contact_derivatives}
\end{eqnarray}
where $\frac{\partial\mathbf{\dot{v}}}{\partial\boldsymbol{\tau}}=\mathbf{M}^{-1} - \mathbf{M}^{-1}\mathbf{J}_\mathcal{C}^\top\widehat{\mathbf{M}}^{-1}\mathbf{J}_\mathcal{C}\mathbf{M}^{-1}$, $\frac{\partial\mathbf{\dot{v}}}{\partial\mathbf{a}_\mathcal{C}}=\mathbf{M}^{-1}\mathbf{J}^\top_\mathcal{C}\widehat{\mathbf{M}}^{-1}$, $\frac{\partial\boldsymbol{\lambda}_\mathcal{C}}{\partial\boldsymbol{\tau}}=-\widehat{\mathbf{M}}^{-1}\mathbf{J}_\mathcal{C}\mathbf{M}^{-1}$ and $\frac{\partial\boldsymbol{\lambda}_\mathcal{C}}{\partial\mathbf{a}_\mathcal{C}}=\widehat{\mathbf{M}}^{-1}$ are the system-acceleration and contact-force Jacobians with respect to inverse dynamics and constrained-acceleration kinematics functions, respectively, $\frac{\partial\boldsymbol{\tau}}{\partial\mathbf{x}}$, $\frac{\partial\boldsymbol{\tau}}{\partial\mathbf{u}}$ are the derivatives of \gls{rnea}, and $\frac{\partial\mathbf{a}_\mathcal{C}}{\partial\mathbf{x}}$, $\frac{\partial\mathbf{a}_\mathcal{C}}{\partial\mathbf{u}}$ are the \gls{fk} derivatives of the computed frame acceleration.
The derivatives of contact dynamics depend on the analytical derivatives of~\gls{rnea}, which favors the computational efficiency as described in~\cite{carpentier-rss18}.

A similar procedure is used to compute the derivatives of the impulse dynamics described in~\eref{eq:kkt_impulse}.
It involves computing the post-impact velocity and contact impulse derivatives as follows:
\begin{eqnarray}\nonumber
\frac{\partial\boldsymbol{\tau}}{\partial\mathbf{w}} &=& \frac{\partial\textrm{RNEA}(\mathbf{q},\mathbf{0},\mathbf{v^+ - v})}{\partial\mathbf{w}} - \frac{\partial\mathbf{g(q)}}{\partial\mathbf{w}},\\
\frac{\partial\mathbf{a}_\mathcal{C}}{\partial\mathbf{w}} &=& e\frac{\partial\textrm{FK}(\mathbf{q},\mathbf{v^+},\mathbf{0})}{\partial\mathbf{w}},
\end{eqnarray}
where $\mathbf{w}$ describes $\mathbf{q}$, $\mathbf{v}$ or $\mathbf{u}$.
Note that partial derivatives of the gravity vector with respect $\mathbf{v}$ and $\mathbf{u}$ are zero.
As before, this procedure also favors computational efficiency, as it depends on the~\gls{rnea} derivatives.

\subsubsection{Friction / wrench cone constraints}
The linearized models of the friction or wrench cone $(\mathbf{C},\mathbf{c})$ are computed from the surface normal $\mathbf{\hat{n}}$ (or more precisely from its rotation matrix), friction coefficient $\mu$, and minimum / maximum normal forces $(\boldsymbol{\underline{\lambda}}, \boldsymbol{\bar{\lambda}})$.
We use an inner representation of the friction cone for more robustness~\cite{caron-icra15}, i.e.,
\begin{equation}
\mathbf{C}\coloneqq \mathbf{A} {}^\mathcal{W}\mathbf{R}_\mathcal{C}^\top,
\end{equation}
where $(\mathbf{A},\mathbf{c})$ are defined as
\begin{eqnarray*}
\mathbf{A}&=&
\left[\begin{matrix}
1 & 0 & -\mu\\
-1 & 0 & -\mu\\
\vdots\\
\cos{\theta_{n_e}} & \sin{\theta_{n_e}} & -\mu\\
-\cos{\theta_{n_e}} & -\sin{\theta_{n_e}} & -\mu\\
0 & 0 & 1
\end{matrix}\right]\in\mathbb{R}^{(n_e+1)\times 3},\\
\mathbf{c}^\top&=&
\left[\begin{matrix}
1 & 1 & \hdots & 1 & \lambda^{\hat{n}}_{\min}
\end{matrix}\right]\in\mathbb{R}^{n_e+1},
\end{eqnarray*}
with ${n_e}$ equals to the number of facets used in the linearized cone, $\theta_{i}$ is the angle of the facets $i$, $\lambda^{\hat{n}}_{\min}\geq 0$ is the minimum unilateral force along the normal \textit{contact} direction $\hat{n}$, and ${}^\mathcal{W}\mathbf{R}_\mathcal{C}\in\mathbb{SO}(3)$ is the cone rotation matrix.
We use the analytical derivatives of the contact forces (or impulses) in~\eref{eq:contact_derivatives} to compute the first-order derivatives of the friction-cone constraints.

In case of six-dimensional contacts, we define a contact wrench cone constraint as described in~\cite{caron-icra15}.
This contact model constrains linear and angular accelerations required for instance for full surface contacts as on bipeds with flat feet (as in~\fref{fig:humanoid_handstand}).
Concretely, it imposes limits for the center of pressure to be inside the support area and yaw torques to be inside a predefined box.
In the interests of space, we omit a complete description, however, it can be expressed in terms of $\mathbf{A}$ and $\mathbf{c}$.
Note that we extend our previous work~\cite{budhiraja-ichr18,mastalli-icra20} by incorporating this friction/wrench-cone constraints.

\subsubsection{Contact placement and velocity constraints}
As the contact placement constraint generally lies on a $\mathbb{SE}(3)$ manifold, we explicitly express it in its tangent space using the $\log(\cdot)$ operator.
Furthermore, the term
\begin{equation}\nonumber
    {}^\mathcal{W}\mathbf{p}_\mathcal{G}^{-1} \cdot {}^\mathcal{W}\mathbf{p}^{ref}_\mathcal{G}
\end{equation}
describes the inverse composition between the reference and current contact placements~\cite{blanco-10se3}, where $^\mathcal{W}\mathbf{p}^{ref}_\mathcal{G}$, ${}^\mathcal{W}\mathbf{p}_\mathcal{G}$ represent the reference and current placements of the set of swing contacts $\mathcal{G}$ expressed in the inertial frame $\mathcal{W}$, respectively.
Note that we compute ${}^\mathcal{W}\mathbf{p}_\mathcal{G}$ using \textit{forward kinematics}.
The results of this operation lies in its tangent space, and it can be interpreted as the difference between both placements, i.e., $^\mathcal{W}\mathbf{p}^{ref}_\mathcal{G}\ominus{}^\mathcal{W}\mathbf{p}_\mathcal{G}$.
We compute the analytical derivatives of the $\log(\cdot)$ operator using \textsc{Pinocchio}~\cite{carpentier-19sii}, however, we suggest the reader to see~\cite{blanco-10se3} for technical details.
We include a contact velocity constraints to define the swing-foot velocity during foothold events.

\subsubsection{Inequality constraints and their resolution}
We enforce the friction-cone, contact placement, and state bounds constraints through quadratic penalization.
For the sake of simplicity, we chose a quadratic penalty function, instead of a standard log-barrier method~\cite{nocedal-optbook}, as it shows good results in practice.
Finally, we handle the control bounds explicitly as explained in~\sref{sec:feas_solver}.

\subsection{Feasibility-driven resolution}\label{sec:feas_solver}
We solve the optimal control problem in \eref{eq:oc_problem} at each~\acrshort{mpc} update using a feasibility-driven \gls{ddp}-type solver which explicitly handles control constraints. %
This method is particularly important for our \gls{mpc} due to three key properties:
\begin{enumerate}
    \item The feasibility-driven search correctly handles dynamic defects (gaps between the roll-out of a policy and the state trajectory) and efficiently converges to feasible solutions for agile motions.
    \item The explicit handling of control constraints, which are also a limiting factor for agile maneuvers, allows our robot to exploit its actuation limits and dynamics while still ensuring fast convergence.
    \item The Riccati recursion applied at each control step provides a local feedback control law. We exploit this property of DDP/iLQR-style methods in \sref{sec:sfc}.
\end{enumerate}

Below, we provide a brief description of our solver named \textsc{Box-F\gls{ddp}}, which we numerically evaluated in \cite{mastalli-underreview20}.

\subsubsection{The Box-FDDP algorithm}\label{sec:boxfddp}
This algorithm solves nonlinear optimal control problems of the form:
\begin{align*}\label{eq:uconstrained_oc}
&{\underset{\mathbf{x}_s,\mathbf{u}_s}{\min}} ~ \ell_N(\mathbf{x}_N) + \sum_{k=0} ^{N-1} \ell_k(\mathbf{x}_k,\mathbf{u}_k) \\\nonumber
&\textrm{s.t.} ~~ \mathbf{x}_{k+1} = \mathbf{f}(\mathbf{x}_k,\mathbf{u}_k),
\hspace{1em}\mathbf{\underline{u}} \leq \mathbf{u}_k + \delta\mathbf{u}_k \leq \mathbf{\bar{u}},
\end{align*}
where $\ell(\cdot)$ describes the terminal or running costs, $\mathbf{f}(\cdot)$ describes the nonlinear dynamics, and $\mathbf{\underline{u}}$,~$\mathbf{\bar{u}}$ are the lower and upper control bounds, respectively.

In similar fashion to the F\gls{ddp} algorithm~\cite{mastalli-icra20}, Box-F\gls{ddp} builds a local approximation of the Hamiltonian $\mathbf{Q}$ while considering the dynamics infeasibility as:
\begin{eqnarray}
    \label{eq:standard_hamiltonian_computation}\nonumber
	\mathbf{Q}_{\mathbf{x}} & = & \boldsymbol{\ell}_{\mathbf{x}} + \mathbf{f}^T_{\mathbf{x}} \mathcal{V}^+_{\mathbf{x}}, \\\nonumber
	\mathbf{Q}_{\mathbf{u}} & = & \boldsymbol{\ell}_{\mathbf{u}} + \mathbf{f}^T_{\mathbf{u}} \mathcal{V}^+_{\mathbf{x}}, \\
	\mathbf{Q}_{\mathbf{xx}} & = & \boldsymbol{\ell}_{\mathbf{xx}} + 
	\mathbf{f}^T_{\mathbf{x}} (\mathcal{V}'_{\mathbf{xx}} + \mu\mathbf{I}) \mathbf{f}_{\mathbf{x}},\\\nonumber
	\mathbf{Q}_{\mathbf{xu}} & = & \boldsymbol{\ell}_{\mathbf{xu}} + 
	\mathbf{f}^T_{\mathbf{x}} (\mathcal{V}'_{\mathbf{xx}} + \mu\mathbf{I}) \mathbf{f}_{\mathbf{u}},\\\nonumber
	\mathbf{Q}_{\mathbf{uu}} & = & \boldsymbol{\ell}_{\mathbf{uu}} + 
	\mathbf{f}^T_{\mathbf{u}} (\mathcal{V}'_{\mathbf{xx}} + \mu\mathbf{I}) \mathbf{f}_{\mathbf{u}} + \mu\mathbf{I},
\end{eqnarray}
where $\boldsymbol{\ell}_{\mathbf{x}}$, $\boldsymbol{\ell}_{\mathbf{u}}$, and $\boldsymbol{\ell}_{\mathbf{xx}}$, $\boldsymbol{\ell}_{\mathbf{xu}}$, $\boldsymbol{\ell}_{\mathbf{uu}}$ are the Jacobians and Hessians of the cost function; $\mathbf{f}_{\mathbf{x}}$, $\mathbf{f}_{\mathbf{u}}$ are the Jacobians of the system dynamics; $\mathcal{V}'_\mathbf{x}$, $\mathcal{V}'_\mathbf{xx}$ are the gradient and Hessian of the Value function with $\mathcal{V}_\mathbf{x}^+\coloneqq\mathcal{V}_\mathbf{x}' + \mathcal{V}_\mathbf{xx}'\mathbf{\bar{f}}$, $\mathbf{\bar{f}}\coloneqq\mathbf{f(x,u)}\ominus{}\mathbf{x}'$ describes the dynamics infeasibility, and $\mu$ is the automatically selected regularization.

This approximation is then used to compute an optimal policy projected in the free space, where the feed-forward control is obtained by solving the following quadratic program:
\begin{equation*}
    \begin{aligned}
        \mathbf{k} &= \arg\min_{\delta\mathbf{u}} \frac{1}{2}\delta\mathbf{u}^\top\mathbf{Q}_{\mathbf{uu}}\delta\mathbf{u} + \mathbf{Q}_{\mathbf{u}}^\top\delta\mathbf{u},\\
        \hspace{-2em}\textrm{s.t.} & \hspace{4em} \mathbf{\underline{u}} \leq \mathbf{u} + \delta\mathbf{u} \leq \mathbf{\bar{u}},
    \end{aligned}
\end{equation*}
and the feedback gain is retrieved as
\begin{equation*}
    \mathbf{K}=-\mathbf{\hat{Q}}^{-1}_{\mathbf{uu},f}\mathbf{Q}_{\mathbf{ux}},
\end{equation*}
where $\mathbf{\hat{Q}}^{-1}_{\mathbf{uu},f}$ is the control Hessian of the free subspace, and this subspace is obtained from the above quadratic program.
For increasing efficiency, we solve this quadratic program using a Projected-Newton QP algorithm~\cite{bertsekas-siam82}.
Finally, the local policy $(\mathbf{k},\mathbf{K})$ is rolled out with a feasibility-driven approach.
To accept a given step length, we use Goldstein condition.

\subsection{Receding horizon and implementation details}
As we know the number of limbs of the robots, the optimization horizon, and set of cost and constraints in advance, we allocate the memory of the predictive controller once.
This reduces the computation time and increases the determinism of the~\acrshort{mpc} loop.
At each computation step of the predictive controller, we update the references of the above-mentioned optimal control problem efficiently and perform a single iteration of the \gls{oc} solver.
We account for expected communication delays between the robot and the predictive controller.
These delays affect the initial state used in the optimal control problem.
Finally, we warm-start the trajectory of states $\mathbf{x}^0_s$, controls $\mathbf{u}^0_s$, and initial regularization $\hat{\mu}^0$ at each~\acrshort{mpc} step.
This regularization value modifies the search direction from Newton to steepest descent and vice versa (for more details see~\sref{sec:feas_solver}).
Below, we provide more details.

\subsubsection{Efficient updates of the \gls{oc} problem}
If the current contact schedule has not been changed by the footstep planner, then, at each~\acrshort{mpc} step, we define the information of the last running models only: the feet positions for the swinging contacts $\mathcal{G}$, and the contact and friction-cone constraints for the active contacts $\mathcal{C}$.
Note that we have to define the last running models as we recede the horizon.
Otherwise, we update the reference of each running model as we wish to generate a new contact schedule.

\subsubsection{Accounting for communication and computation delays}
As the predictive controller runs on a dedicated computer, it receives the information of the current state of the robot within a predictable delay.
To predict the changes in the robot state, we run a forward simulation, i.e.,
\begin{equation*}
\mathbf{x}^{pred}_0 = \mathbf{f}_{p_0}(\mathbf{x}_0, \mathbf{u}_0, \Delta t^{delay}),
\end{equation*}
where $\mathbf{f}_{p_0}(\cdot)$ is the system dynamics given a current set of contacts $p_0$, $\Delta t^{delay}$ is the expected communication and computation delay, $\mathbf{x}_0$ and $\mathbf{u}_0$ are the current state and control at the published time, and $\mathbf{x}^{pred}_0$ is the initial (predicted) state used by the~\acrshort{mpc}.
With this approach, we assume that the current joint torques $\mathbf{u}_0$ are applied in the delay interval, which it is reasonable for short delay periods ($\sim$\SI{2.5}{\milli\second}).

\subsubsection{Warm-starting and regularization}
We use \textit{direct-indirect hybridization} (we coined the term in our work~\cite{mastalli-underreview20}) to formulate the optimal control problem.
It means that we can initialize both: state and control trajectories.
Thus, when we start the predictive controller, we define the state and control guesses for all the nodes as the robot's nominal posture (i.e.,  $\mathbf{x}^{nom}\coloneqq[\mathbf{q}^{nom},\mathbf{0}]$) and the quasi-static torque commands needed to maintain this posture $\mathbf{u}^{qstatic}$, respectively.
We compute the quasi-static torques as
\begin{equation}\label{eq:qstatic_torques}
\begin{bmatrix}
\mathbf{u}^{qstatic} & \boldsymbol{\lambda}^{qstatic}
\end{bmatrix} = 
\begin{bmatrix}
\mathbf{S} & \mathbf{J}^\top_\mathcal{C}(\mathbf{q}^{nom})
\end{bmatrix}^{+}\mathbf{g}(\mathbf{q}^{nom}),
\end{equation}
where $()^+$ is the pseudo-inverse, and $\boldsymbol{\lambda}^{qstatic}$ if the contact forces for the quasi-static condition.
Otherwise, we use the previous solution to warm-start the common running nodes between the current and next~\acrshort{mpc} updates.
Furthermore, as in the startup procedure, we use the nominal posture and quasi-static torques for the last running nodes, i.e., the nodes that we include with the receded horizon.
We avoid unnecessary online computations by storing the results computed in~\eref{eq:qstatic_torques}.

The initial regularization provides information about how to compute the search direction.
Larger regularization values compute a steepest descent direction, instead values close to zero compute a Newton direction.
Therefore, it is recommended to use a large value if we are far from a local minima.
However, as described above, our solver iteratively evaluates this condition and adapts its value, which could increase or decrease.
We use this updated regularization value to warm-start the next step of the~\acrshort{mpc}.
This approach matches the numerical evolution expected when we solve the problem completely.

%% file: src/4_control.tex
\section{Control Policy}\label{sec:control_policy}
In this section, we present two methods for building a control policy from the outputs of a full-dynamics~\acrshort{mpc}.
First, we describe an instantaneous \gls{wbc} based on inverse dynamics and hierarchical quadratic programming (\sref{sec:wbc}).
This motion-planning approach is broadly used to control quadruped (e.g.,~\cite{bellicoso-ichr16,shamelmastalli-ral19}) and humanoid robots (e.g.,~\cite{herzog-ar16,delpret-ras15}), but it leads to sub-optimal policies.
Therefore, in~\sref{sec:sfc}, we introduce a novel \gls{sfc} controller leveraging the Riccati gains computed by the \gls{mpc}.
This controller builds a local optimal policy from first principles of optimization (cf.~\sref{sec:feas_solver}).
It constitutes what we refer to as motion-control paradigm.
Finally, we include details on state estimation and drift compensation in~\sref{sec:state_estimation}.

\subsection{Whole-body control}\label{sec:wbc}
Typically instantaneous control actions (or joint torque commands) aim to track reference tasks while ensuring the physical realism both in terms of dynamics, balancing and joint limits.
Common reference tasks in legged robotics are \gls{com} and swing-foot motions, body orientation, angular momentum, and contact force.
This problem is formalized as a quadratic program, which can incorporate slack variables to allow temporal violations of the constraints (e.g.,~\cite{herzog-iros14,shamelmastalli-ral19}).
We can solve it with general-purpose quadratic optimization software as in~\cite{delprete-tro16,shamelmastalli-ral19}, or we can employ null-space projectors to form hierarchical quadratic programs that encode task priorities as in~\cite{herzog-ar16,xin-icra18}.
Below, we briefly describe our whole-body control approach that is formulated as a hierarchical quadratic program.
It extends previous work~\cite{xin-icra18} by incorporating momentum and contact forces tasks.
\aref{sec:hierarchical_wbc} explains its generic formulation and resolution for readers that are not familiar with whole-body control based on a hierarchy of quadratic programs.

\subsubsection{System dynamics}
We aim to track reference \gls{com} and swing-foot motions, centroidal momentum, and contact forces computed from our full-dynamics~\acrshort{mpc}.
To find these instantaneous torque commands, we define the highest priority to the contact dynamics (physical realism), friction cone (balancing), and torque limits (actuation).
This means that our first quadratic program is described as
\begin{equation}\label{eq:qp1}
\begin{aligned}
& \underset{\mathbf{y}}{\min}
& & \frac{1}{2}\left\|\begin{bmatrix} \mathbf{M} & -\mathbf{S} & -\mathbf{J}_\mathcal{C}^\top\\\mathbf{J}_\mathcal{C} & \mathbf{0} & \mathbf{0}\end{bmatrix}\mathbf{y}-\begin{bmatrix}-\mathbf{h} \\-\mathbf{\dot{J}}_\mathcal{C}\mathbf{v}\end{bmatrix}\right\|_2^2 \\
& \text{s.t.}
&& \begin{bmatrix}\mathbf{0} & \mathbf{0} & \mathbf{C}\end{bmatrix}\mathbf{y} \leq \mathbf{0}, \\
&&& \mathbf{\underline{u}}\leq\begin{bmatrix}\mathbf{0} & \mathbf{I} & \mathbf{0}\end{bmatrix}\mathbf{y}\leq\mathbf{\bar{u}},
\end{aligned} 
\end{equation}
where $\mathbf{y}\coloneqq[\mathbf{\dot{v}}\,\,\mathbf{u}\,\,\boldsymbol{\lambda}]$ defines the decision variables, $(\mathbf{M}, \mathbf{h}, \mathbf{S})$ describe the robots' full dynamics, $(\mathbf{J}_\mathcal{C}, \mathbf{\dot{J}}_\mathcal{C})$ denotes the holonomic contact constraint, $\mathbf{C}$ denotes the friction or wrench cone with unilateral constraints, and $(\mathbf{\underline{u}},\mathbf{\bar{u}})$ are the joint torque limits.
For more details about these terms see~\sref{sec:oc_formulation}.

\subsubsection{Motion policy}
When we formulate a cascaded task in our whole-body control formulation, we observe from~\eref{eq:hqp} in \aref{sec:hierarchical_wbc} that the inequality constraints defined in~\eref{eq:qp1} remain the same for each subsequent quadratic program.
Indeed, the \textit{trajectory-tracking} quadratic program changes the quadratic cost only:
\begin{equation}
\begin{aligned}
& \underset{\mathbf{y}}{\min}
& & \frac{1}{2}\left\|\begin{bmatrix}\mathbf{J}_\mathcal{E} & \mathbf{0} & \mathbf{0}\end{bmatrix}\mathbf{y}+\mathbf{\dot{J}}_\mathcal{E}\mathbf{v}-\mathbf{\dot{v}}_{\mathcal{E},\mathrm{d}}\right\|_2^2 \\
& \text{s.t.}
& & \begin{bmatrix}\mathbf{0} & \mathbf{0} & \mathbf{C}\end{bmatrix}\mathbf{y} \leq \mathbf{0}, \\
&&& \mathbf{\underline{u}}\leq\begin{bmatrix}\mathbf{0} & \mathbf{I} & \mathbf{0}\end{bmatrix}\mathbf{y}\leq\mathbf{\bar{u}}.
\end{aligned} 
\end{equation}
This ensures the kinematic feasibility as $\mathbf{J}_\mathcal{E}\mathbf{\dot{v}}+\mathbf{\dot{J}}_\mathcal{E}\mathbf{v}=\mathbf{\dot{v}}_{\mathcal{E},\mathrm{d}}$, where
$\mathbf{J}_\mathcal{E}$ is the task Jacobian and the \textit{user-tuned motion policy} $\mathbf{\dot{v}}_{\mathcal{E},\mathrm{d}}$ is described as
\begin{equation*}
\mathbf{\ddot{x}}_\mathrm{d}+\mathbf{K}_e(\mathbf{x}_\mathrm{d}-\mathbf{x})+\mathbf{D}_e(\mathbf{\dot{x}}_\mathrm{d}-\mathbf{\dot{x}})
\end{equation*}
with $(\mathbf{K}_e,\mathbf{D}_e)$ gains that map errors on trajectory tracking into acceleration commands, $\mathbf{\ddot{x}}_\mathrm{d}$ is a feed-forward reference acceleration, and $\mathbf{x}_\mathrm{d}$,  $\mathbf{\dot{x}}_\mathrm{d}$ are the reference position and velocity of the task.
We use this formulation to describe \gls{com} and swing-foot tasks computed by the~\acrshort{mpc}, in which the swing-foot tracking has a higher priority.

\subsubsection{Momentum policy}
By employing the centroidal momentum matrix $\mathbf{A}_G$ proposed in~\cite{orin-ar13,wensing-ijhr16}, we formulate the following quadratic program to track the reference centroidal momentum.
\begin{equation}
    \begin{aligned}
        & \underset{\mathbf{y}}{\text{min}}
        & & \frac{1}{2}\left\|\begin{bmatrix}\mathbf{A}_{G} & \mathbf{0} & \mathbf{0} \end{bmatrix}\mathbf{y}+\mathbf{\dot{A}}_G\mathbf{v}-\mathbf{\dot{h}}_{G, \,\mathrm{d}}\right\|_2^2 \\
        & \text{s.t.}
& & \begin{bmatrix}\mathbf{0} & \mathbf{0} & \mathbf{C}\end{bmatrix}\mathbf{y} \leq \mathbf{0}, \\
&&& \mathbf{\underline{u}}\leq\begin{bmatrix}\mathbf{0} & \mathbf{I} & \mathbf{0}\end{bmatrix}\mathbf{y}\leq\mathbf{\bar{u}},
    \end{aligned} 
\end{equation}
where $\mathbf{\dot{h}}_{G,\, \mathrm{d}} = \begin{matrix}[\mathbf{\dot{k}}_{G,\, \mathrm{c}}^\top & \mathbf{\dot{l}}_{G,\, \mathrm{c}}^\top]\end{matrix}^\top\in\mathbb{R}^6$ is the reference rate of change in the centroidal momentum composed of linear component and angular component.
This \textit{user-tuned momentum policy} is defined as follows
\begin{eqnarray}\nonumber\label{eq:momentum_policy}
&&\mathbf{\dot{l}}_{G, \, \mathrm{c}}=m[\mathbf{\ddot{r}}_{G, \, \mathrm{d}}+\mathbf{K}_l(\mathbf{r}_{\mathrm{G, \, \mathrm{d}}}-\mathbf{r}_G)+\mathbf{D}_l(\mathbf{\dot{r}}_{G, \, \mathrm{d}}-\mathbf{\dot{r}}_G)],\\
&&\mathbf{\dot{k}}_{G, \, \mathrm{c}}=\mathbf{\dot{k}}_{G, \, \mathrm{d}}+\mathbf{D}_{k}(\mathbf{k}_{G, \, \mathrm{d}}-\mathbf{k}_G),
\end{eqnarray}
where $\mathbf{r}_{G,\, \mathrm{d}}\in\mathbb{R}^3$ denotes the reference~\gls{com} position computed by the~\acrshort{mpc}, $\mathbf{r}_G\in\mathbb{R}^3$ is the actual~\gls{com} position observed by state estimators, $m$ denotes the total mass of the robot, $\mathbf{k}_{G,\mathrm{d}}\in\mathbb{R}^3$ and $\mathbf{k}_G\in\mathbb{R}^3$ are the reference and actual angular momentum, respectively, $\mathbf{K}_l$ and $\mathbf{D}_i \, (i=l, \, k) $ are feedback gains.

\subsubsection{Contact-force policy}
After satisfying all the above tasks, we use the redundancy to track the reference contact forces $\boldsymbol{\lambda}_\mathrm{d}\in\mathbb{R}^{n_f}$ generated by the predictive controller when there are two or more feet in contact
\begin{equation}
\begin{aligned}
& \underset{\mathbf{y}}{\min}
& & \frac{1}{2}\left\|\begin{bmatrix} \mathbf{0} & \mathbf{0} & \mathbf{I}\end{bmatrix}\mathbf{y} - \boldsymbol{\lambda}_\mathrm{d}\right\|_2^2 \\
& \text{s.t.}
&& \begin{bmatrix}\mathbf{0} & \mathbf{0} & \mathbf{C}\end{bmatrix}\mathbf{y} \leq \mathbf{0}, \\
&&& \mathbf{\underline{u}}\leq\begin{bmatrix}\mathbf{0} & \mathbf{I} & \mathbf{0}\end{bmatrix}\mathbf{y}\leq\mathbf{\bar{u}}.
\end{aligned}\label{eq:contact_policy}
\end{equation}
This task is important for dynamic gaits as contact forces play an important role in the momentum generation.
This last quadratic program defines the \textit{contact-force policy}.

\subsubsection{Flight phase control}\label{sec:wbc_flight_phase_control}
With any whole-body controller, unfortunately, we need to treat flight phases (i.e., where no legs are in contact) as a special case.
It is required since these type of controllers are highly reliant on good state estimation, and when the robot is flying, we can no longer integrate odometry from leg kinematics (cf.~\sref{sec:state_estimation}).
To mitigate these significant inaccuracies in the state estimation, we rely on a PD controller with feed-forward torques computed by the~\acrshort{mpc}.

\subsection{Riccati-based state-feedback control}\label{sec:sfc}
\begin{figure}[!tb]
    \centering
    \includegraphics[width=0.95\columnwidth]{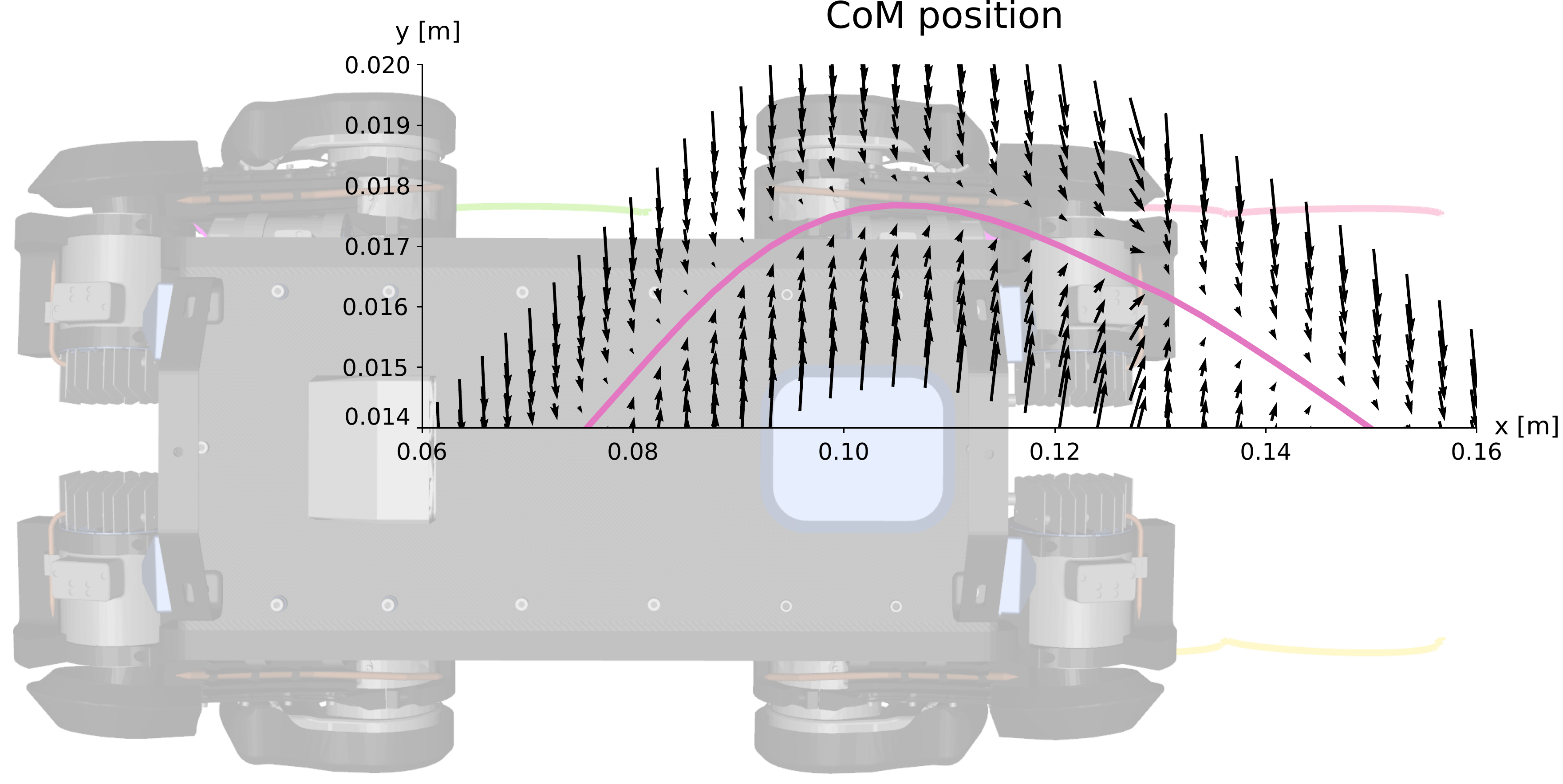}
    \caption{Optimal policy for a forward trotting gait computed by our predictive controller.
    As shown in the plot, the optimal policy changes locally given the current state of the system.
    For sake of simplicity, we display the policy on the \gls{com} motion, however, our predictive controller computes optimal policies for the whole-body motion.
    For visualization, we use different scales for the \textit{\gls{com} policy} and ANYmal robot.}
    \label{fig:policy_flow}
\end{figure}

As briefly discussed above, the classical whole-body controller tracks a set of individual user-tuned tasks: \gls{com}, swing-foot, angular momentum, and contact force tasks.
Often, these user-tuned tasks \textit{fill the gaps} left by using reduced-order dynamics in the motion generation (e.g., inverted pendulum~\cite{mastalli-tro20} or single-rigid body dynamics~\cite{aceituno_cabezas-ral18,melon-icra21}).
However, instead of hand-tuning a set of sub-optimal policies that aim to track an optimal trajectory, we could compute both optimal whole-body motions and their local policies if we plan motions using the full-body dynamics:
\begin{equation}\label{eq:optimal_policy}
\mathbf{u}^\mathrm{d} = \boldsymbol{\pi}(\texttt{model}, \mathbf{x}) = -\overbrace{\mathbf{\hat{Q}}^{-1}_{\mathbf{uu},f}\mathbf{Q_u}}^{\mathbf{u}^*_{ff}} - \overbrace{\mathbf{\hat{Q}}^{-1}_{\mathbf{uu},f}\mathbf{Q_{ux}}}^{\mathbf{K}}(\mathbf{x}^*\ominus\mathbf{x}),
\end{equation}
where $\mathbf{\hat{Q}}_{\mathbf{uu},f}$, $\mathbf{Q_{ux}}$ are the local approximation of the Hamiltonian as described in~\sref{sec:boxfddp}.
This local and optimal policy around the optimal state $\mathbf{x}^*$ is defined by an optimal control problem $(\texttt{model})$ that considers cost, dynamics and constraints.
Indeed, it corresponds to the optimal cost-to-go $\mathcal{V}(\mathbf{x})$ as depicted in~\fref{fig:policy_flow}.

Since our optimal control solver (cf.~\sref{sec:feas_solver}) computes the state-feedback gain matrix $\mathbf{K}$ as a byproduct of its backward pass, we can build a \textit{state-feedback} controller that strictly follows the optimal policy $\boldsymbol{\pi}\in\mathbb{R}^{n_u}$.
This helps to achieve variable impedance based on the task priorities defined in the predictive controller and to guarantee torque limits, as the Riccati state-feedback matrix $\mathbf{K}$ couples feedback actions between the different tasks and projects them to the free control space.
As thus, control limits are handled implicitly.
Another advantage of this approach is that there is no need for additional tuning as in the whole-body controller, which will unfortunately \textit{rewrite} the policy.
Furthermore, it is computationally inexpensive when compared with solving several quadratic programs during each control tick at runtime (as described in~\sref{sec:wbc}).
In practice, computing the policy takes less than \SI{100}{\micro\second} on our limited on-board hardware---this compares with \SIrange{1}{2}{\milli\second} for the whole-body controller.
However, to make execution of the computed policy work safely and reliably, several crucial modifications have to be made as described below.

Our Riccati state-feedback controller differs from~\cite{grandia-iros19}.
In this works, the authors proposed a feedback method that maps errors in the state into contact force commands, rather than joint torque commands.
Then, the system is forward simulated under such feedback term, which computes desired accelerations and eventually reference torques via a whole-body controller.
This control architecture inherits the same limitations of a motion-planning paradigm, i.e., it does not follow the optimal policy.

\subsubsection{Updating reference state from a system rollout}
We interpolate the \acrshort{mpc} references while guaranteeing the physical realism, as these references are often sampled at a different discretization frequency compared with the control frequency (here, commonly, \SI{100}{\hertz} and \SI{400}{\hertz}, respectively).
Hereto, we run a \textit{forward prediction} of the reference state $\mathbf{x}^*$, through a system rollout, to compute the feedback term in~\eref{eq:optimal_policy}.
With the \textit{forward-predicted} state, we then obtain the applied joint torques $\mathbf{u}_\mathrm{d}$ as well as desired joint position $\mathbf{q}_{j,\mathrm{d}}$ and velocities $\mathbf{v}_{j,\mathrm{d}}$ sent to the low-level joint-torque controller~\cite{hutter-advro-2017}.
This forward-prediction and interpolation step produces smoother joint torques, which is crucial for executing the policy on real hardware.

\subsubsection{Handling state estimation delays}
Delays and inaccuracies in the base state estimation affects the execution of control policy $\boldsymbol{\pi}$, as its feedback action closes the loop on base pose and twist.
Similarly to the whole-body controller during flight phases (cf.~\sref{sec:wbc_flight_phase_control}), we modify the state-feedback gain $\mathbf{K}$ when the base estimate is unreliable.
Concretely, we deactivate the components of feedback gain associated to the base error signals when less than two feet are in contact as leg-odometry corrections are inaccurate~\cite{camurri-ral17}.
It consists of setting the base terms to zero, i.e.,
\begin{equation*}
  \mathbf{K} = \left[\begin{matrix} \mathbf{0} & \mathbf{K}_q^j & \mathbf{0} & \mathbf{K}_v^j\end{matrix}\right],
\end{equation*}
where $\mathbf{K}^j_q$, $\mathbf{K}^j_v\in\mathbb{R}^{n_j\times n_v}$ are the feedback gains associated to the joint positions and velocities, respectively.
This holistic approach allows to keep tracking of desired joint states closely to the optimal policy, which is mainly driven by the feed-forward torque $\mathbf{u}^*_{ff}$.
Additionally, it adds robustness and compliance against inaccurate contact state estimates~\cite{camurri-ral17} as, again, the system is driven by the optimal feed-forward torque and joint-feedback actions following closely the policy $\boldsymbol{\pi}$.

\begin{figure}[b]
    \includegraphics[width=\columnwidth]{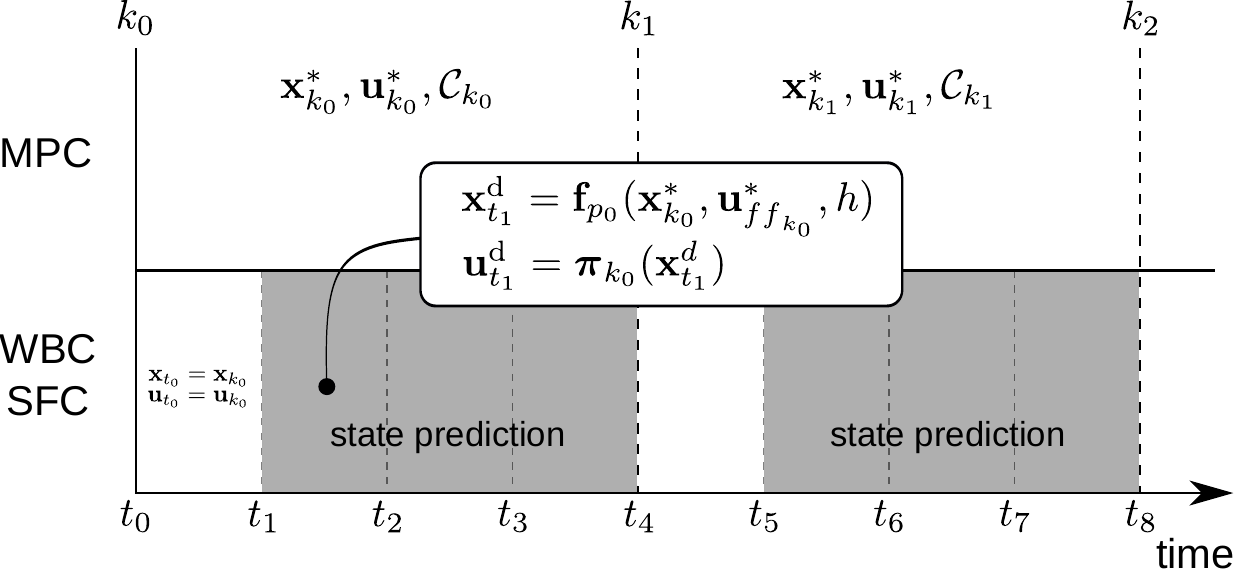}
    \caption{Illustration of the state prediction between \acrshort{mpc} and control discretization points.
    In the gray areas, we predict the reference states $\mathbf{x}^\mathrm{d}_{t_1}$ using the contact dynamics $\mathbf{f}_{p_0}$, which are subject to a desired set of active contacts $\mathcal{C}_{p_0}$.
    This state is then used to compute a local control policy $\boldsymbol{\pi}_{k_0}$ either from the \acrfull{wbc} or \acrfull{sfc}.
    Note that $\{k_0,k_1,\cdots\}$ describe the~\gls{mpc} discretization nodes and $\{t_0,t_1,\cdots\}$ are the time instances where we compute a new local policy.}
    \label{fig:mpc_forward_state_prediction}
\end{figure}

\subsection{Forward state prediction from system rollout}
As mentioned earlier, while waiting for a new \acrshort{mpc} update, we wish to interpolate the reference state at the control-policy frequency.
To do so, we perform a contact-consistent forward simulation using the \textit{feed-forward torque} $\mathbf{u}^*_{ff}$.
This computes the predicted (reference) states within the \acrshort{mpc} period $\{\mathbf{x}^\mathrm{d}_1, \cdots, \mathbf{x}^\mathrm{d}_{i+1}\}$:
\begin{eqnarray}
\mathbf{x}^\mathrm{d}_1 &=& \mathbf{f}_p(\mathbf{x}^{*}, \mathbf{u}^*_{ff}, h), \nonumber\\
\mathbf{x}^\mathrm{d}_2 &=& \mathbf{f}_p(\mathbf{x}^\mathrm{d}_1, \mathbf{u}^*_{ff}, h), \nonumber\\
&\vdots& \nonumber\\
\mathbf{x}^\mathrm{d}_{i+1} &=& \mathbf{f}_p(\mathbf{x}^\mathrm{d}_i, \mathbf{u}^*_{ff}, h),
\end{eqnarray}
where $h$ is the control period, $\mathbf{x}^\mathrm{d}_i$ is the predicted state at controller time $i$, the \acrshort{mpc} period is equals to $\Delta t = (i+1)h$, $\mathbf{f}_p(\cdot)$ defines the contact dynamics which are subject to a set of desired active contacts $\mathcal{C}_p$ at phase $p$.
We also consider the reflected inertia of the actuators in the contact dynamics.
We illustrate this concept in~\fref{fig:mpc_forward_state_prediction}.

\subsubsection{State prediction in both controllers}
We apply the state prediction procedure in both the whole-body and state-feedback controllers, as they run at higher frequencies compared with the~\gls{mpc} loop.
In the state-feedback controller, we use the predicted state $\mathbf{x}^\mathrm{d}_i$ to compute the error in the state $\mathbf{x}^\mathrm{d}_i\ominus\mathbf{x}_i$, where $\mathbf{x}_i$ is the current measurement of the robot's state.
In the whole-body controller, $\mathbf{x}^\mathrm{d}_i$ is used to compute the references of the~\gls{com} and swing-foot, momentum, and contact forces tasks.

\subsection{State estimation and drift compensation}\label{sec:state_estimation}
We estimate the base pose and twist by fusing inertial sensing from the onboard IMU, and kinematic and contact information using an~\gls{ekf}. 
It computes both the base pose ${}^\mathcal{O}\mathbf{\hat{M}}_\mathcal{B}\in\mathbb{SE}(3)$ and twist ${}^\mathcal{O}\mathbf{\hat{v}}_\mathcal{B}\in\mathbb{R}^6$ with respect to the \textit{odometry frame} $\mathcal{O}$.
We use the state estimator provided on the ANYmal robot (TSIF, \cite{bloesch-rss12}).

The above state estimator provides continuous and smooth estimates, making it ideal for use in a closed-loop controller.
However, it accumulates drift due to inaccuracies in the kinematics and contact interaction (e.g., point contact assumption ignores the geometry of the foot or compressibility of the foot or terrain).
We correct this drift by matching successive LiDAR scans with the iterative closest point algorithm~\cite{pomerleau-auro-13}.
This provides the relative transform ${}^\mathcal{W}\mathbf{\hat{M}}_\mathcal{O}$ between an \textit{inertial frame} $\mathcal{W}$, which has limited drift but discontinuous updates, and the odometry frame $\mathcal{O}$.

We can compensate the drift at either the~\gls{mpc} or controller level.
In the former case, we first transform the current robot measurements expressed in the odometry frame into the inertial one.
Then, we transform the planned motion expressed in the inertial frame into the odometry one.
In the latter case, the~\gls{mpc} plans entirely in the inertial frame, while the controller transforms plans to the odometry frame upon receipt.
In practice, both cases perform equally well.

%% file: src/5_results.tex
\section{Results} \label{sec:results}
In this section, we show experimental trials on the generation of agile and complex maneuvers on ANYmal robots.
We also show the benefits of following the optimal policy by comparing our Riccati-based state-feedback controller against a state-of-the-art instantaneous controller.
All the experimental results are shown in the
accompanying video or in \texttt{\url{https://youtu.be/R8lFti7x5N8}}.

\subsection{Experimental setup}
We evaluate the performance of our hybrid~\gls{mpc} using two different ANYmal robots.
Both ANYmal robots are series B~\cite{hutter-advro-2017}.
These quadrupeds have a nominal base height of \SI{\approx50}{\centi\metre}, weigh \SI{\approx30}{\kilo\gram}, and their twelve series-elastic actuators are capable of achieving peak torques of \SI{40}{\newton\metre}.
The robots are equipped with depth and LiDAR sensors for terrain perception and localization and mapping, respectively (\emph{A:} \textsc{Multisense SL-7} and rotating \textsc{Hokuyo}, \emph{B:} \textsc{Realsense D435} and \textsc{Velodyne VLP-16}).

The state estimator and controller run on the onboard PC (2 cores i7-7500U CPU @ \SI{2.70}{\giga\hertz}), with the LiDAR localization running a separate identical onboard computer.
The~\gls{mpc} runs on an offboard computer connected via \textsc{Ethernet} (\emph{A:} 8 cores i9-9980HK CPU @ \SI{2.4}{\giga\hertz}, \emph{B:} 6 cores i7-10875H CPU @ \SI{2.3}{\giga\hertz}).
The onboard and offboard computers communicate via a low-latency ROS communication layer (TCP, no-delay).
Both offboard~\gls{mpc} computers are capable of computing policy updates over a horizon of \SI{1.25}{\second} at a rate in excess of \SI{100}{\hertz}.
For determinism, we throttle the \acrshort{mpc} rate to \SI{50}{\hertz}.

\begin{figure}[b]
    \includegraphics[width=\columnwidth]{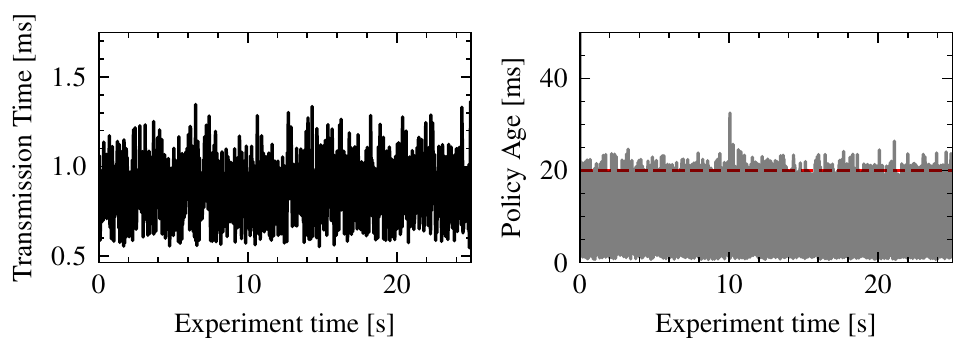}
    \caption{Policy transmission time and policy age for a stair climbing experiment. %
    \emph{Left:} Time for serialization from Python, communication via Ethernet, and deserialization of the policy between the computer running the~\gls{mpc} and the control computer. The transmission time is consistently less than one control tick.
    \emph{Right:} Age of the executed policy/time since the last~\gls{mpc} update.
    The red line indicates the theoretical maximum age for an update rate of \SI{50}{\hertz} without considering roundtrip communication delay.}
    \label{fig:policy_lag}
\end{figure}

\subsection{Computation frequency and communication delay}
First, we investigate the policy lag and age for PC setup \emph{B}.
In~\fref{fig:policy_lag}, we can see that the transmission time from the control computer remains below one control tick and that the whole-body control policy updates are steady at \SI{50}{\hertz}.
In our \acrshort{mpc} pipeline, the policy transmission time is the time needed to encode it from \textsc{NumPy}/\textsc{Python} to \textsc{Eigen}/\textsc{C++}, to send it to via \textsc{Ethernet} and \textsc{ROS-TCP/IP}, and to decode it upon receipt.
On average, updating the \gls{mpc} problem takes \SIrange{1}{2}{\milli\second} using \textsc{Python}, and solving the updated problem takes between \SIrange{10}{12}{\milli\second}.
This computation time depends on the number of backward or forward passes performed by our \textsc{Box-FDDP} solver.

\subsection{Variable impedance under state-feedback control}
\begin{figure}[t]
    \centering\begin{tabular}{cc}
    \rowname{a} & \href{https://youtu.be/R8lFti7x5N8?t=141}{\raisebox{-.5\height}{\includegraphics[width=0.9\columnwidth]{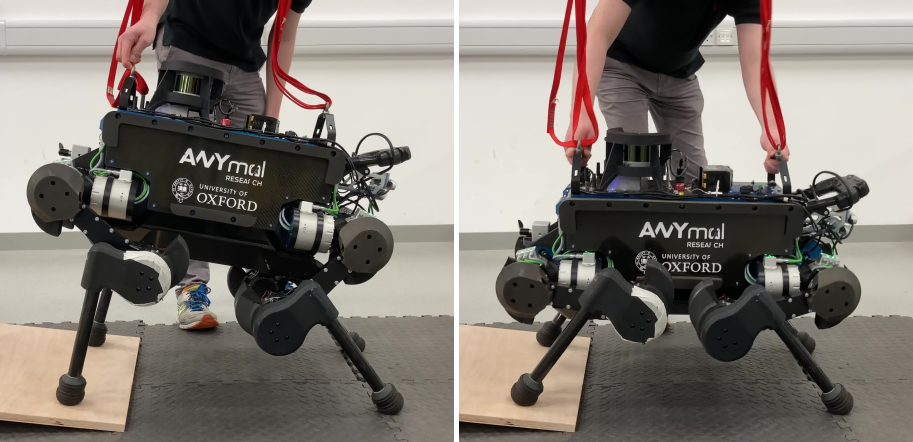}}}\\
    & \rule[0.2em]{8cm}{0.1pt}\\
    \rowname{b} & \href{https://youtu.be/R8lFti7x5N8?t=141}{\raisebox{-.5\height}{\includegraphics[width=0.9\columnwidth]{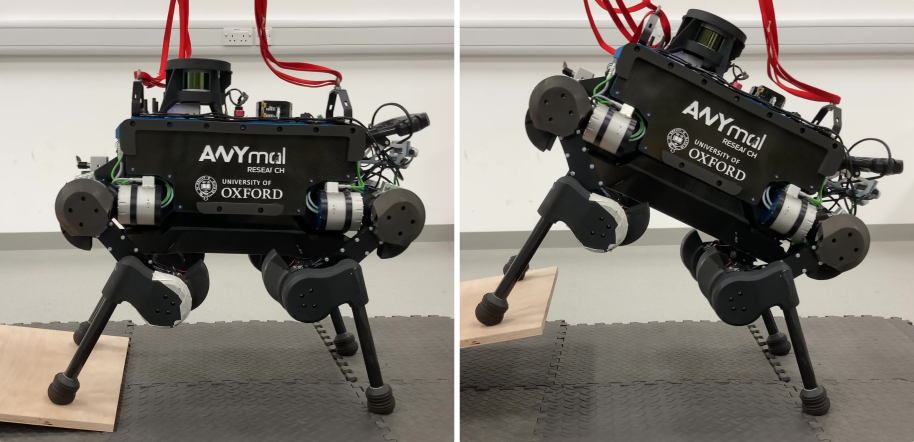}}}
    \end{tabular}
    \caption{ANYmal balancing and following a~\gls{com} reference using state-feedback~\gls{mpc} with on-board computation.
    (a) the Riccati-based state-feedback together with the \textit{high-frequency} predictive control show variable compliance through soft and strong interaction with the robot.
    (b) the control pipeline is able to quickly stabilize contact perturbations.
    For this experiment, the~\gls{mpc} policy with time horizon \SI{0.4}{\second} is recomputed onboard at \SI{90}{\hertz}.
    To watch the video, click its respective figure or see \texttt{\url{https://youtu.be/R8lFti7x5N8}}}
    \label{fig:compliance_control_exp}
\end{figure}

We validated the ANYmal's stability and compliance using our Riccati-based state-feedback controller.
This controller automatically incorporates, from \textit{first-principles of optimization}, compliance while guaranteeing stability and feasibility as defined in the~\gls{mpc} problem.
\fref{fig:compliance_control_exp} shows high compliance and stability that occurred when our~\acrshort{mpc} tracked a fixed~\gls{com} reference.
In the first part of the experiment, we pushed the ANYmal robot towards the ground.
It moved compliantly while still guaranteeing the torque limits and stability, which is very restrictive closer to the ground~\cite{orsolino-ral18,orsolino-tro20}.
In the second part of the experiment, we moved the ramp supporting the hind feet (balance board).
Here, our~\gls{mpc} controller adjusted the trunk orientation and maintained the contact stability.
This strategy helps to reduce the torque limits while keeping contact forces closely aligned with the gravitational field.
The time horizon of the~\gls{mpc} was \SI{0.4}{\second} and we run it onboard at \SI{90}{\hertz}.
Similarly, we have also explored the robustness of our predictive control approach for balancing to different contact properties, sudden disturbances (kicking), and walking gaits with unmodeled payloads. We included these qualitative evaluations in our accompanying video.

\begin{figure*}%
    \centering\begin{tabular}{cc}
    \rowname{a} & \href{https://youtu.be/R8lFti7x5N8?t=167}{\raisebox{-.5\height}{\includegraphics[width=0.93\textwidth]{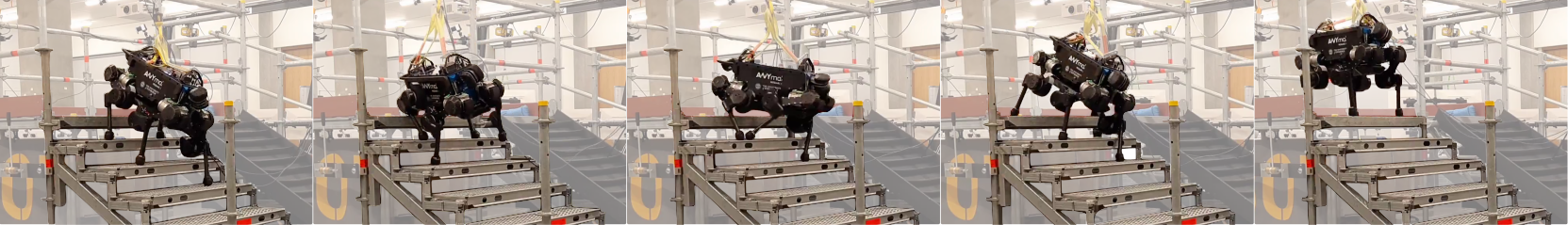}}}\\
    & \rule[0.5em]{16.9cm}{0.1pt}\\
    \rowname{b} & \href{https://youtu.be/R8lFti7x5N8?t=236}{\raisebox{-.5\height}{\includegraphics[width=0.93\textwidth]{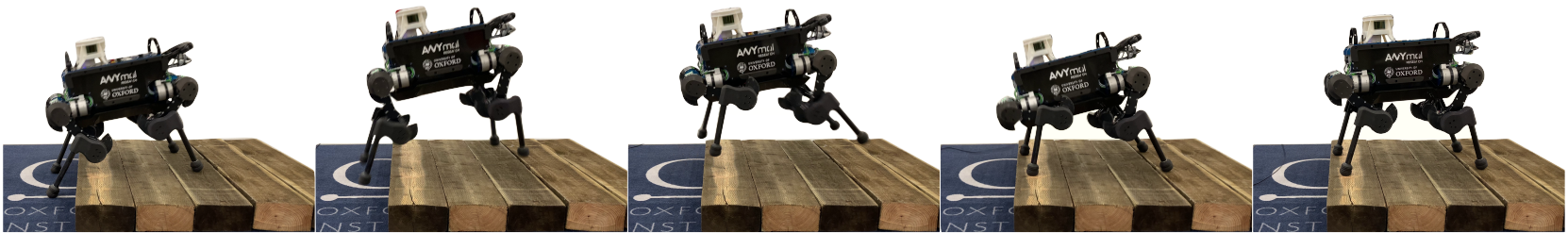}}}\\
    & \rule[0.5em]{16.9cm}{0.1pt}\\
    \rowname{c} & \href{https://youtu.be/R8lFti7x5N8?t=103}{\raisebox{-.5\height}{\includegraphics[width=0.93\textwidth]{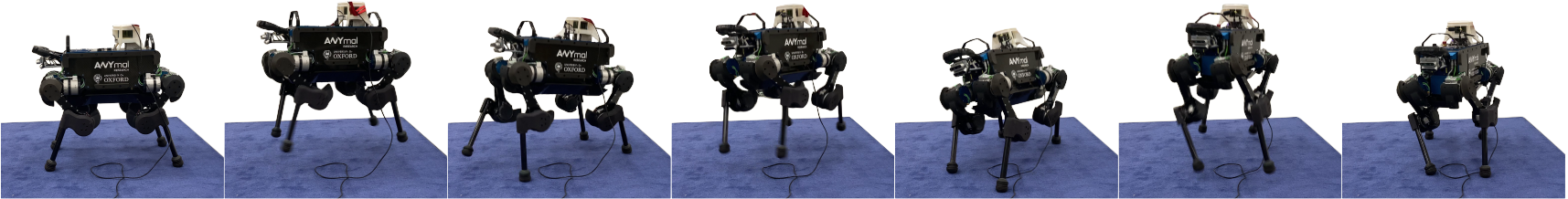}}}\\
    & \rule[0.5em]{16.9cm}{0.1pt}\\
    \rowname{d} & \href{https://youtu.be/R8lFti7x5N8?t=224}{\raisebox{-.5\height}{\includegraphics[width=0.93\textwidth]{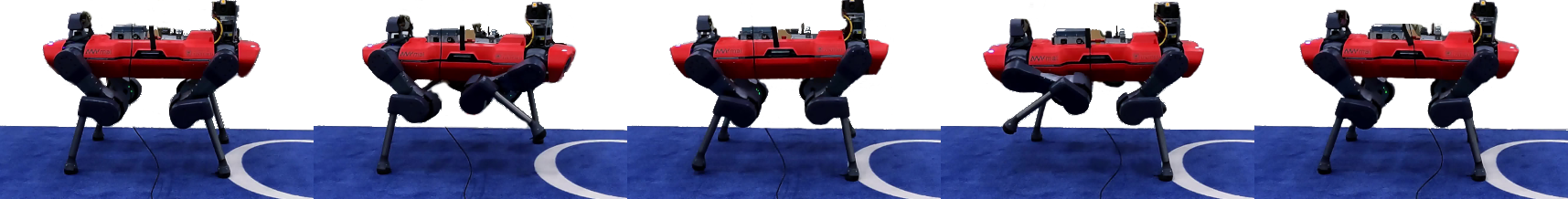}}}%
    \end{tabular}
    \caption{Snapshots of different locomotion maneuvers computed by our hybrid model predictive control.
    All these experiments trials use the Riccati-based state-feedback controller described in Section \ref{sec:sfc} and onboard state estimation based on proprioceptive and exteroceptive sources.
    (a) climbing up industrial stairs with~\SI{28}{\centi\metre} of length and~\SI{17}{\centi\metre} of height.
    (b) multiple forward jumps of~\SI{20}{\centi\metre} over a pallet of~\SI{12}{\centi\metre} of height.
    (c) multiple twist jumps of 30 degrees each one.
    (d) transfer of our MPC framework to the ANYmal C quadruped (\SI{50}{\kilo\gram}).
    To watch the video, click its respective figure or see \texttt{\url{https://youtu.be/R8lFti7x5N8}}.}
    \label{fig:highly_dyn_maneuvers}
\end{figure*}

\subsection{Agile maneuvers with full-dynamics MPC}
We tested our hybrid~\gls{mpc} and Riccati-based state-feedback controller in different agile maneuvers.
\fref{fig:highly_dyn_maneuvers} shows snapshots of relevant instances in which our ANYmal B robot performed industrial-like climbing stairs, forward jumps over a pallet, and rotating jumps on the spot.
To the best knowledge of the authors, our method is the first to generate multiple jumps continuously on the ANYmal robot, which were possible only with our Riccati controller and when we considered the robot's torque limits in our \acrshort{mpc}.
To assert the benefits of using full-body dynamics and tracking the optimal policy, we planned the footstep locations for the entire motion in each of these experiments once.
We imposed accurate torque limits of our ANYmal B, which are \SI{40}{\newton\metre}, discretized the trajectory every \SI{10}{\milli\second} (\SI{100}{\hertz}), re-planned the maneuver for a receding horizon of \SI{1.25}{\second}, and run the~\gls{mpc} and Riccati controller loops at \SI{50}{\hertz} and \SI{400}{\hertz}, respectively.
Note that in the accompanying video we report more experimental trials.

\begin{figure}
    \centering\begin{tabular}{cr}\vspace{1em}
    \rowname{a} & \raisebox{-.5\height}{\href{https://youtu.be/R8lFti7x5N8?t=189}{\includegraphics[width=0.92\columnwidth]{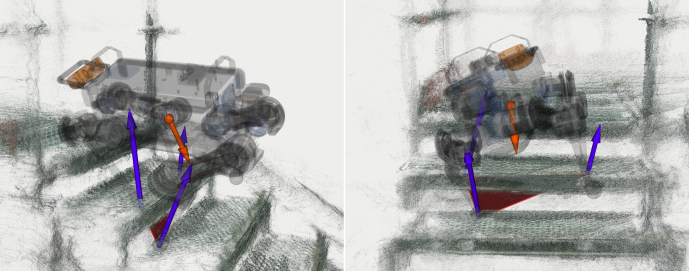}}}\\
    \rowname{b} & \raisebox{-.5\height}{\includegraphics[width=0.92\columnwidth]{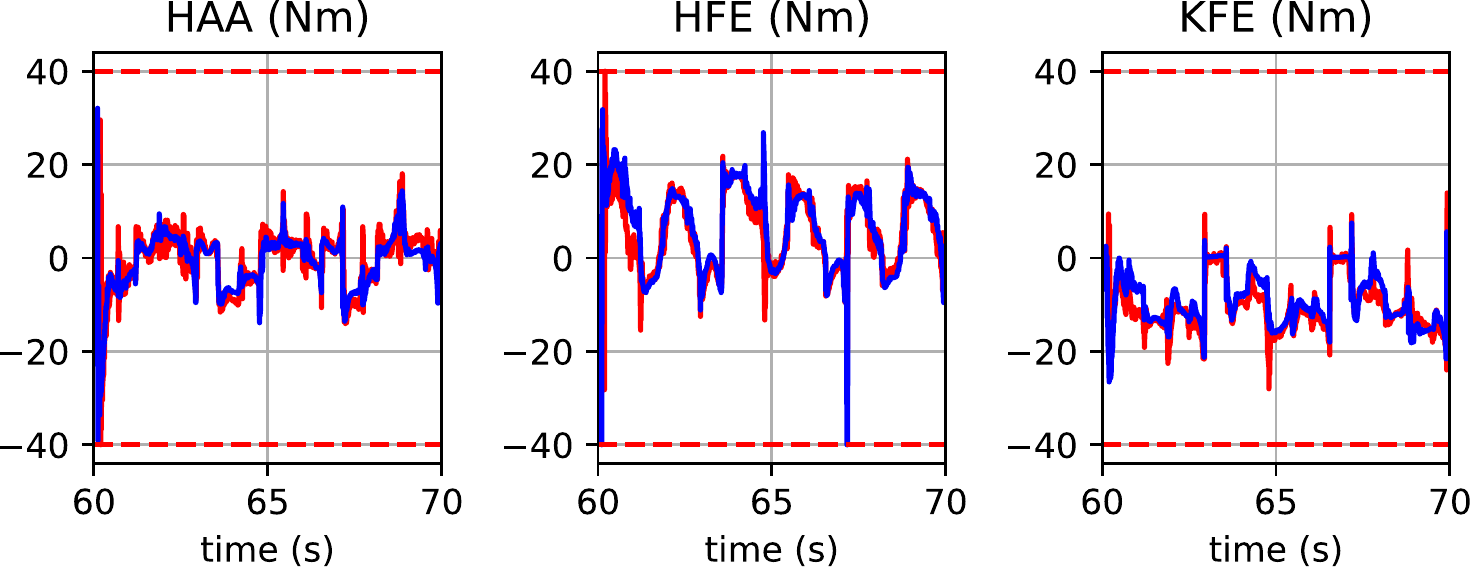}}
    \end{tabular}
    \caption{Posture and torque optimization for climbing stairs.
    (top) Support polygon (red), contact forces (purple) and~\gls{com} position and velocity (orange) in an unexpected moment when the ANYmal robot was climbing the stairs.
    In the second picture of~\fref{fig:highly_dyn_maneuvers}, we can see the actual robot.
    The small support polygon is due to an unexpected slippage of the LF foot, which was triggered by inaccuracies in the state estimation.
    (bottom) Range of joint torques generated during the stairs climbing experiments.
    The reference torques computed by the state-feedback controller are in blue, while the measured torques are in red.}
    \label{fig:climbing_up}
\end{figure}

\subsubsection{Climbing up industrial stairs}
Handling torque limits is a key element to climb up industrial stairs with our ANYmal B robot.
We built a scaffolding stair with 7 treads, which resembles the stairs typically use in construction sites.
Its riser height and tread length are \SI{17}{\centi\metre} and \SI{26}{\centi\metre}, respectively, which forms an inclination of \SI{33}{\degree}.
This climbing task is challenging for the ANYmal B robot due to its limited actuation (i.e., torque limits).
Indeed, our robot cannot climb up this stairs with the built-in controllers provided by ANYbotics~\cite{bellicoso-ichr16,fankhauser-icra18}.
\fref{fig:highly_dyn_maneuvers}a displays the sequence of maneuvers optimized in real-time by our~\gls{mpc}.
This complex maneuver was generated reactively after the robot missed three steps due to state estimation inaccuracies.
Furthermore, \fref{fig:climbing_up}a shows the optimized posture needed to reduce the torque commands in the presence of a very small support region.
In all the stairs-climbing experiments, we observed that the torque commands are usually below to \SI{25}{\newton\meter} as shown in~\fref{fig:climbing_up}b.
Finally, to achieve this posture, it was also important to track the optimal policy, as it generated the optimal momenta.
For more details see results reported in~\sref{sec:sfc_and_wbc_comparison}.

\subsubsection{Forward jumps onto pallet}
We defined a sequence of footholds and timings that describe a set of multiple forward jumps onto a pallet with \SI{12}{\centi\metre} of height. 
Our full-dynamics~\acrshort{mpc} generated and executed a total of five jumps on the ANYmal B robot.
Each jump has a length of \SI{20}{\centi\metre} and swing-foot height of \SI{12}{\centi\metre}.
Each flight and jumping phases are \SI{280}{\milli\second} and \SI{450}{\milli\second}, respectively.
The \acrshort{mpc} horizon is \SI{1.25}{\second}, which is longer than a single jumping gait.
\fref{fig:highly_dyn_maneuvers}b depicts snapshots of this experiment.
The Riccati state-feedback controller allowed our predictive controller to make rapid changes in torso orientation.
These rolling motions help to account for inaccuracies in the state estimation that happened during the flight phases.
These inaccuracies are in the order of $10$ to \SI{18}{\centi\meter} per jump in vertical axis, if we fuse the odometry of leg only.
It means that our predictive controller had to handle these inaccuracies in some jumps, as the correction from LiDAR scans occurred at \SI{1}{\hertz}.
We performed other experiments on multiple jumps as reported in the accompanying.
In those experiments, our classical whole-body controller destabilized the robot after a few jumps (typically two).
These are the limitations that we have observed when a controller does not track the \acrshort{mpc} policy.

\begin{figure}[b]
    \centering
    \href{https://youtu.be/R8lFti7x5N8?t=29}{\includegraphics[width=\columnwidth]{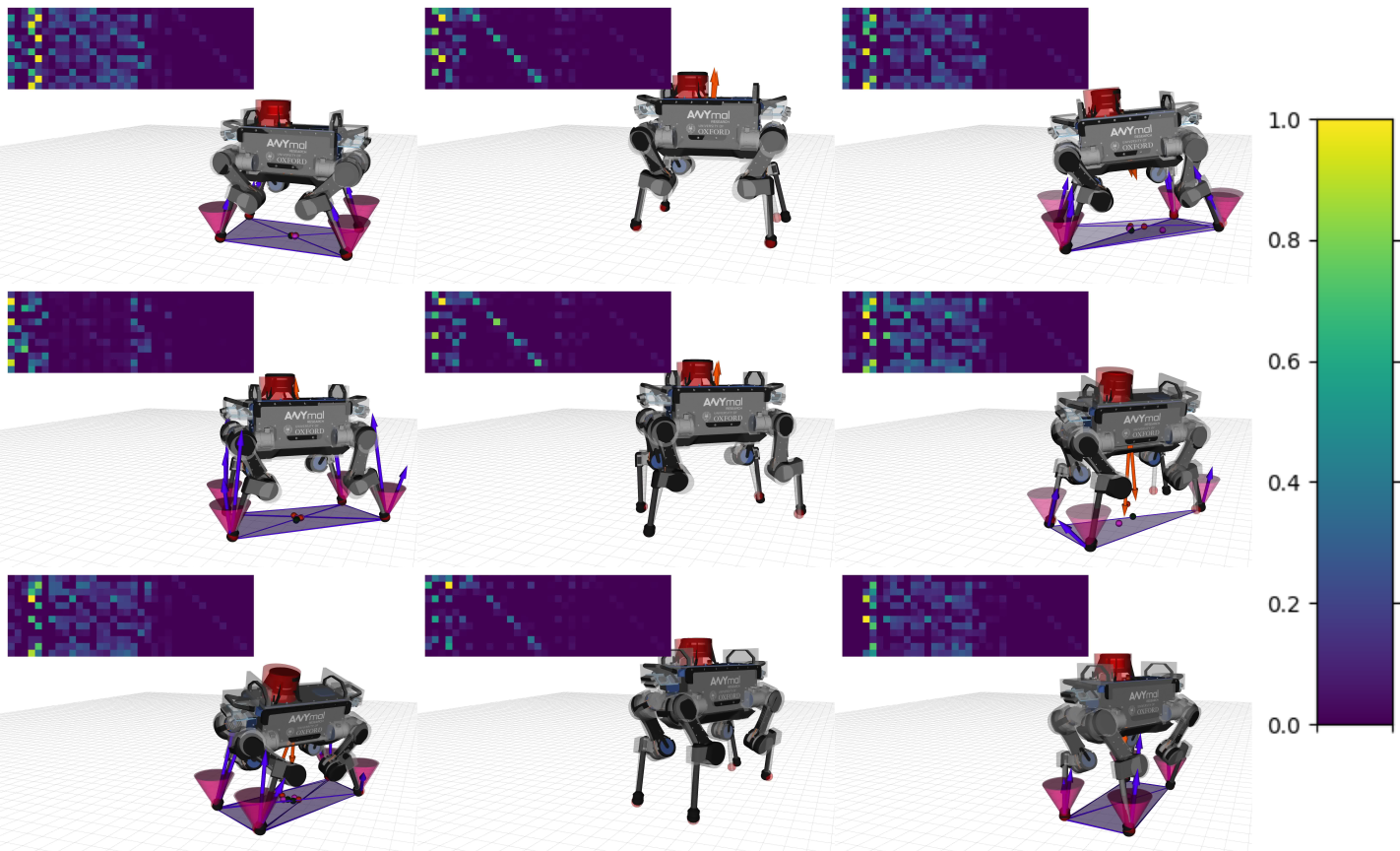}}
    \caption{From left to right, top row to bottom row: Visualization of desired (gray) and measured (colored) states along with the feedback gain matrix for the twist jump experiment shown in \fref{fig:highly_dyn_maneuvers}c.
    The feedback gain matrices used in the state-feedback controller change automatically during the motion execution and recover inconsistent contact timings and leg slippage.
    We display the normalized gaits using a the colormap described in the colorbar.}
    \label{fig:twist_jumps_feedback_gains}
\end{figure}

\subsubsection{Twist jumps in place}
We conducted further experimental trials on jumping motions.
Here, we defined the footholds and timings of multiple jumps that twisted the ANYmal robot by \SI{30}{\degree} each.
We used the footstep timings, \acrshort{mpc} horizon and jumping height defined in previous experiment (jumping onto pallet).
\fref{fig:highly_dyn_maneuvers}c shows the sequence of highly-dynamic motions performed by the ANYmal robot.
This sequence consisted of three jumps that rotated the robot by a total of \SI{90}{\degree}.
In the first jump, we observed a high-impact landing instant that produced an unexpected loss of contact in the~\gls{rf} foot.
As before, this abrupt landing is due to inaccuracies in the state estimation.
However, in the next jump, our predictive and Riccati controllers adjusted the ANYmal's torso roll orientation to account for a shorter \textit{take-off} phase.
This adaptation can also be seen in the changing feedback gain matrices shown in \fref{fig:twist_jumps_feedback_gains}.
During the flight phases (middle column), feedback for the joint tracking and base-joint correlation are dominant.
Instead, stronger correlation between all joints can be observed during the support phases.
It is important to note that multiple jumps posed significant challenges in the motion generation, as each jumps is highly influenced by the performance of the previous one.
In other words, landing appropriately leads to an effective subsequent jump.

\begin{figure}
    \centering\begin{tabular}{cr}\vspace{1em}
    \rowname{a} & \raisebox{-.5\height}{\href{https://youtu.be/R8lFti7x5N8?t=52}{\includegraphics[width=0.92\columnwidth]{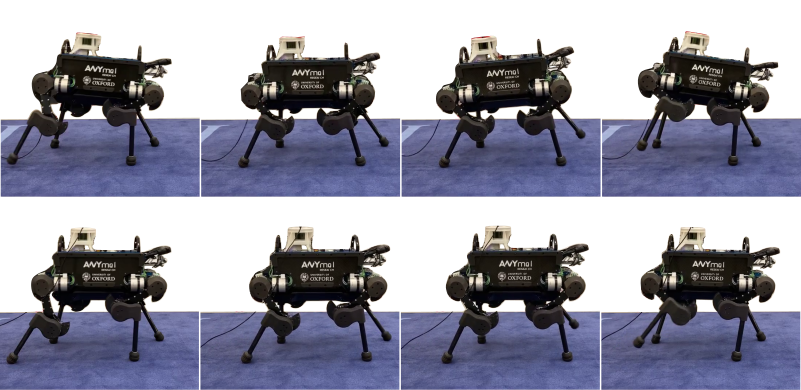}}}\\
    \rowname{b} & \raisebox{-.5\height}{\includegraphics[width=0.92\linewidth]{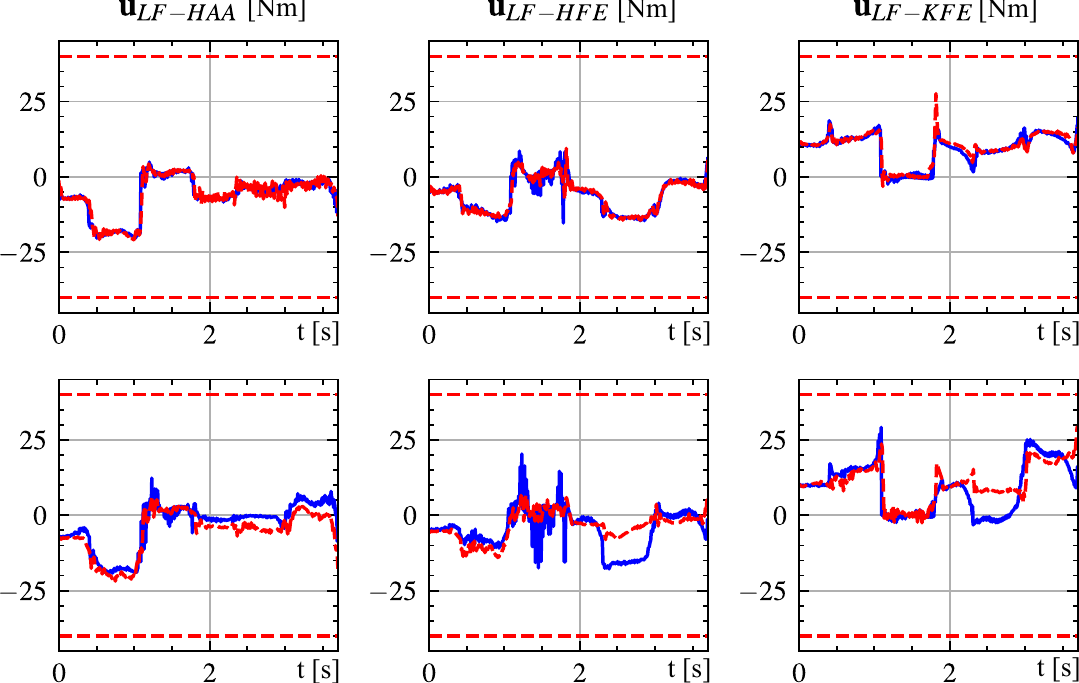}}
    \end{tabular}
    \caption{Walking behavior and torque tracking comparison between the state-feedback controller (top row) and whole-body controller (bottom row).
    The state-feedback controller increased the body rotation (or angular momenta), while whole-body controller kept the body straight.
    Increasing the angular momenta may seem counter-intuitive, however, it reduced the knee (KFE) torque commands.
    Blue are reference/desired and red are measured torques.
    Note that the \gls{wbc} frequently commands torque references that cannot be tracked (e.g., for the KFE joints), whereas the position tracking is good with both controllers.
    To watch the video, click its respective figure or see \texttt{\url{https://youtu.be/R8lFti7x5N8}}.}
    \label{fig:sfc_and_wbc_walk}
\end{figure}

\subsection{Comparison between state-feedback controller and whole-body controller}\label{sec:sfc_and_wbc_comparison}
\begin{figure}[t]
    \centering\begin{tabular}{cr}\vspace{1em}
    \rowname{a} & \raisebox{-.5\height}{\includegraphics[width=0.8925\linewidth]{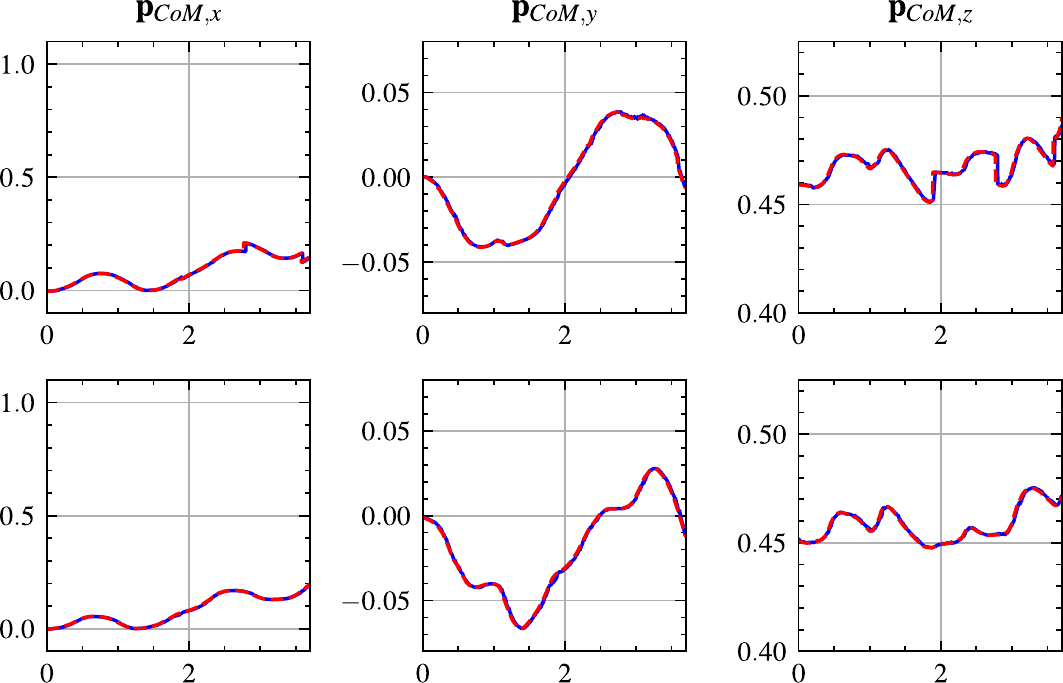}}\\\vspace{1em}
    \rowname{b} & \raisebox{-.5\height}{\includegraphics[width=0.92\linewidth]{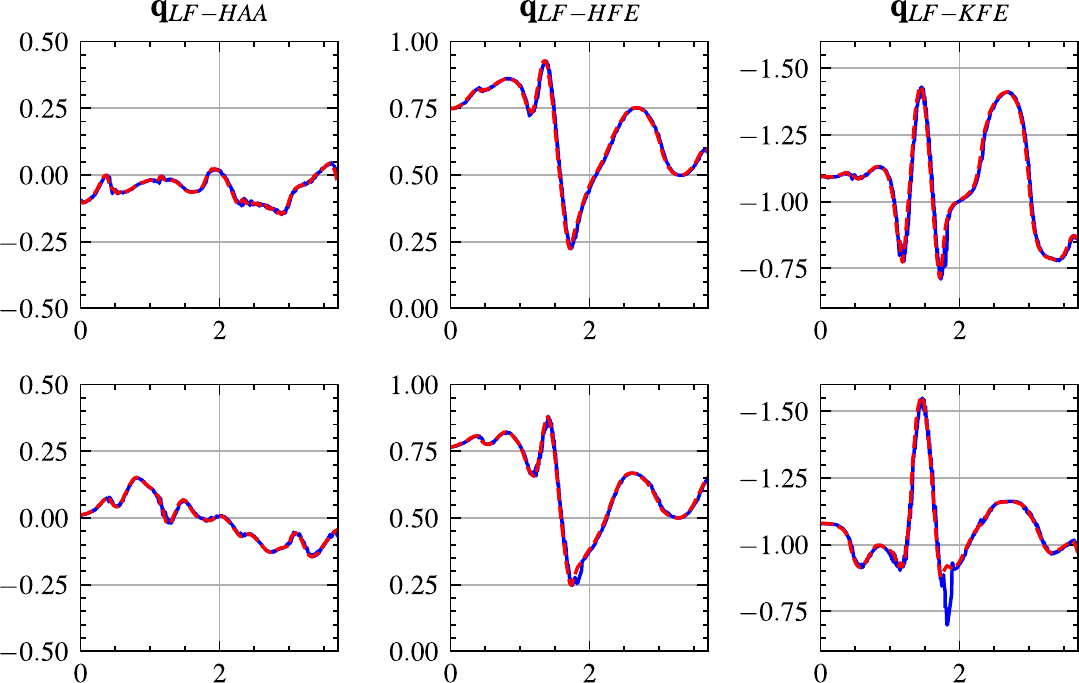}}\\%\vspace{1em}
    \rowname{c} & \raisebox{-.5\height}{\includegraphics[width=0.8925\linewidth]{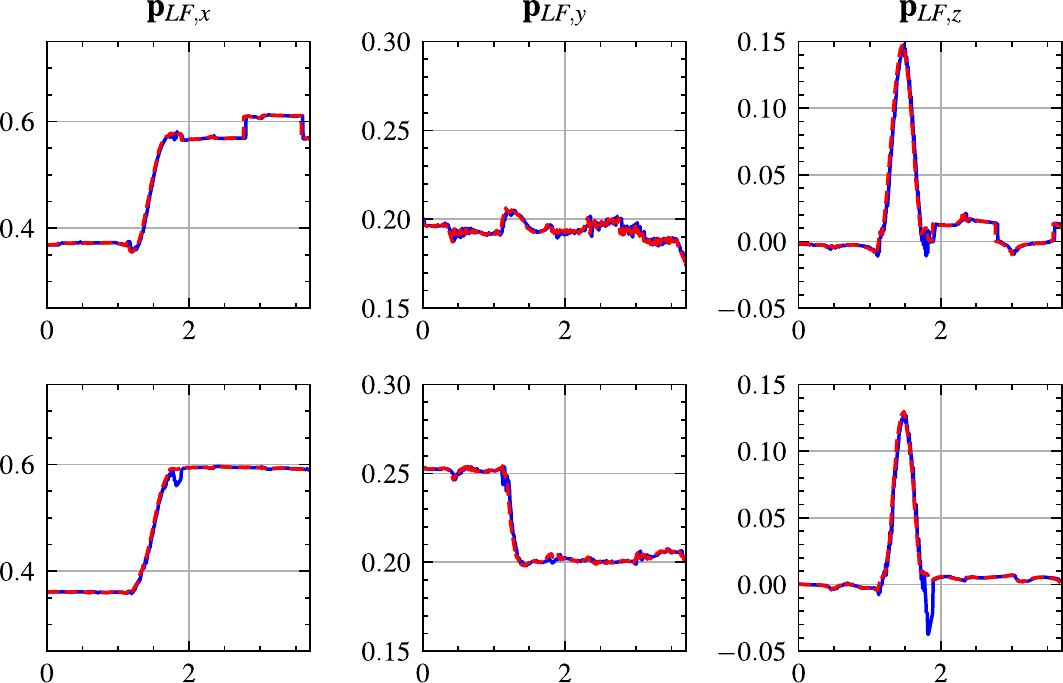}}
    \end{tabular}
    \caption{\gls{com}, joint and contact position tracking comparison between state-feedback controller (top row) and whole-body controller (bottom row) for a walk gait cycle.
    Blue are reference/desired and red are measured quantities.
    The $x$-axis shows time in seconds and the $y$-axes are shown in \si{\metre}, \si{\radian} and \si{\metre} for the \gls{com}, joint and swing plots, respectively.
    Both controllers tracked accurately these quantities.
    However, the \gls{sfc} achieved the desired target step height of \SI{15}{\centi\metre}, whereas the \gls{wbc} cannot achieve the task requirement defined in the \acrshort{mpc}.
    This latter one also deviated significantly from its lateral target position of \SI{20}{\centi\metre}.}
    \label{fig:com_joint_swing_tracking}
\end{figure}

\begin{figure}[t]
    \centering\begin{tabular}{cr}\vspace{1em}
    \rowname{a} &  \raisebox{-.5\height}{\includegraphics[width=0.92\linewidth]{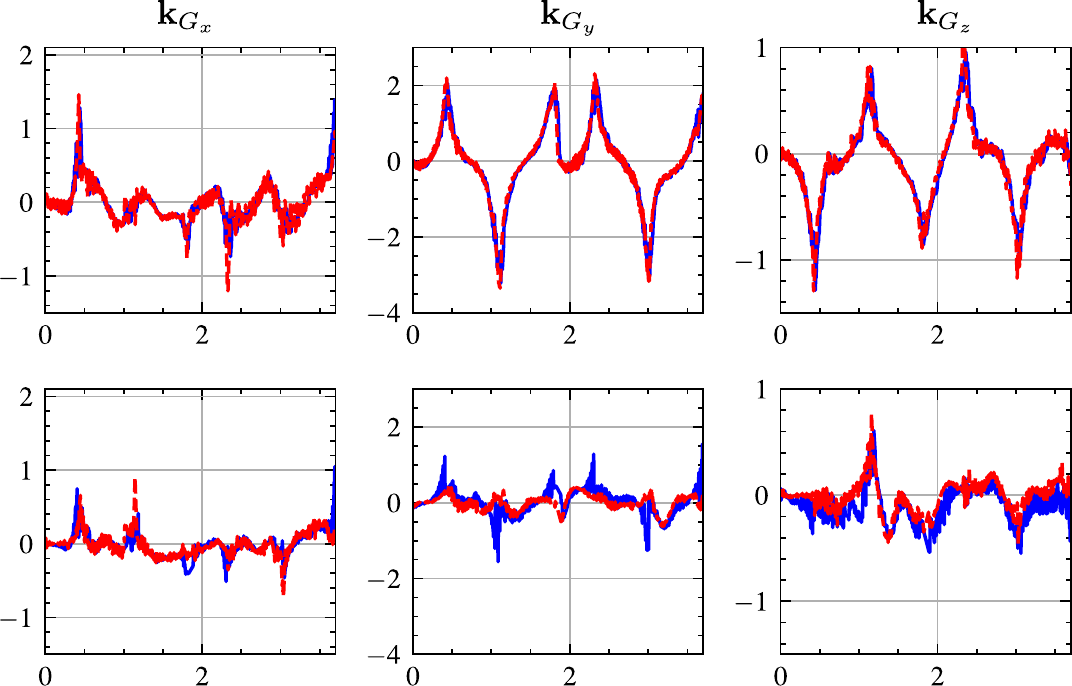}}\\
    \rowname{b} &  \raisebox{-.5\height}{\includegraphics[width=0.92\linewidth]{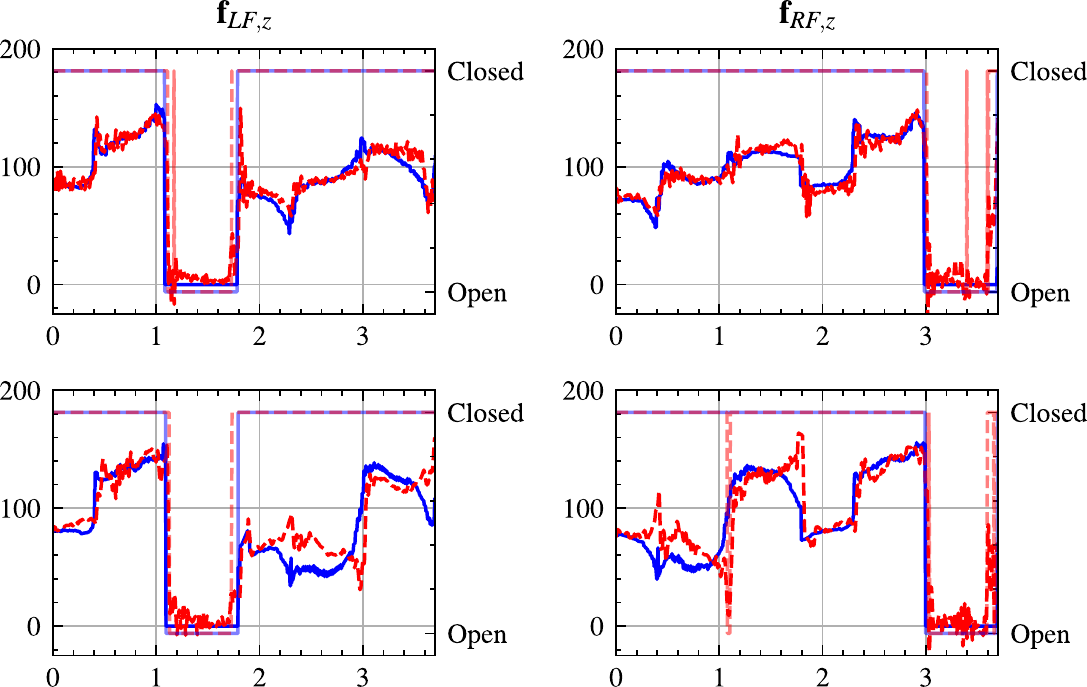}}
    \end{tabular}
    \caption{Momenta and contact force tracking comparison between state-feedback controller (top row) and whole-body controller (bottom row) for a walk gait cycle.
    Blue are reference/desired and red are measured values.
    The x-axis shows time in seconds and the y-axis shows the corresponding SI unit for the quantity.
    The \gls{sfc} closely tracks the desired momenta---even if large---whereas the additional optimization in the \gls{wbc} alters the desired references from the \acrshort{mpc}.
    For visualization, we filtered the estimated contact forces using an 8th-order Butterworth low-pass filter with a cut-off frequency of \SI{80}{\hertz}, since they are susceptible to joint velocity and inertial acceleration noise.}
    \label{fig:momenta_force_tracking}
\end{figure}

We compared our Riccati-based state-feedback control against the classical whole-body control on a walking gait with step height of \SI{15}{\centi\metre} (\fref{fig:sfc_and_wbc_walk}a).
Both controllers tracked the \gls{com}, joint and swing motions accurately, as reported in~\fref{fig:com_joint_swing_tracking}.
However, the classical whole-body controller cannot track properly the torque commands (\fref{fig:sfc_and_wbc_walk}b).
The quality in the torque tracking is related to the accuracy in the generation of the optimal momenta.
This is due to the nonholonomy nature of angular momenta, which cannot be properly tracked with a \textit{time-invariant momentum policy} (used in whole-body controllers).
Indeed, \fref{fig:momenta_force_tracking}a provided experimental evidence of this fact, since only the state-feedback controller tracked and tracks the angular momenta properly.
The tracking inaccuracies in the angular momentum introduced by the whole-body controller translated, and often translates in practice, into higher torque commands (\fref{fig:sfc_and_wbc_walk}b), larger contact forces (\fref{fig:momenta_force_tracking}b), and task deviations (\fref{fig:com_joint_swing_tracking}c).
In \fref{fig:com_joint_swing_tracking}c, we show that the whole-body controller could not achieve the desired target step height of \SI{15}{\centi\metre} and deviated from its lateral target position of \SI{20}{\centi\metre} defined in the \acrshort{mpc}.
Note that (i) to cope for the errors in the momentum tracking, the \acrshort{mpc} changes its reference position and velocity, which still leads to an accurate swing tracking, and (ii) we used the same \acrshort{mpc} formulation and weights for both experiments.
However, these changes are not enough to achieve the desired step height and lateral motion.
Despite that these tracking errors are not critical for quadrupedal walking, they destabilize our robot during multiple jumps.

We obtained these results with a state-of-the-art and well-tuned whole-body controller.
It incorporates a momentum policy proposed by~\cite{orin-ar13,wensing-icra13} and integrates a contact-force policy that improves the torque and momentum tracking considerably.
Note that these aspects increased importantly the tracking performance of the whole-body controller.
In our experiments, we used diagonal matrices for the proportional and damping gains in each task defined in the whole-body controller.
These diagonal values are $60$ and $90$ for proportional and damping gains, respectively.
Instead, for the swing task, they are $350$ and $25$.
Note that similar issues (i.e., higher torque commands,
larger contact forces, and task deviations) appear with different gains.

%% file: src/6_conclusion.tex
\section{Conclusion}\label{sec:conclusion}
In this work, we focused on the generation of agile and complex maneuvers in legged robots.
We started by describing the importance of considering the nonholonomic aspects of the kinetic momenta and robot's actuation limits, which are neglected in state-of-the-art approaches due to its numerical optimization complexity.
We showed that predictive control can nicely leverage these challenges by synthesizing motions and control policies within the actuation limits.
This is possible by employing the robot's full or centroidal dynamics.
However, we explained that using the centroidal dynamics model does not provide a computational benefit in predictive control but rather a disadvantage, as it requires to re-compute the manipulator dynamics for retrieving the joint torques and developed advanced algorithms to handle these nonlinear inequality constraints.

To enable agile maneuvers in legged robots, we proposed a hybrid model predictive control approach that relies on contact and switch dynamics (i.e., robot's full dynamics).
We extended this formulation by incorporating friction/wrench cone constraints via the analytical derivatives of the contact and impulse forces.
Our feasibility-driven formulation and resolution both converge quickly enough for generating these complex maneuvers in real-time.
Our approach handles the actuation limits efficiently, computes locally optimal feedback policies, and is easily transferred to other robots, such as the ANYmal C quadruped.
As far as we know, we are the first to achieve these aspects.
Furthermore, we proposed a Riccati-based state-feedback controller that follows the optimal policy computed by the \gls{mpc}.
We showed that our controller properly generates the angular momentum, which justifies the importance of a motion-control paradigm and \acrshort{mpc} with full-body dynamics.
This is in contrast to whole-body controllers, which fundamentally cannot track the angular momentum accurately needed for generating multiple jumps.
We demonstrated that, in practice, the motion-planning architecture increases the joint torques and contact forces.
It also affects significantly the robot's stability and its capabilities during the execution of agile maneuvers, and indeed, the ANYmal robot can only execute multiple jumps continuously when it follows the optimal policy computed by our~\acrshort{mpc} (i.e., via our Riccati-based state-feedback controller).
We provided theoretical and experimental evidence that show the benefits of our approach in a series of motions of progressively increasing complexity.
Our approach enabled the ANYmal B robot to execute multiple jumps and to reduce the torques needed to climb stairs.
To the best of our knowledge, our work is the first to generate agile maneuvers such as rotating jumps, multiple jumps or jumps over a pallet on the ANYmal robot, which has heavier legs and lower control bandwidth (due to its series elastic actuators) than small quadrupeds such as MIT's Mini Cheetah~\cite{bledt-iros18} and Solo robots~\cite{grimminger-ral20}.

%% file: src/7_appendices.tex
\section{Hierarchical whole-body control}\label{sec:hierarchical_wbc}
We can impose each task of whole-body controller by solving a hierarchy of quadratic programs.
Indeed, the following generic quadratic program can be used to describe each task:
\begin{equation}
\begin{aligned}
& \underset{\mathbf{y}_i}{\min}
& & \frac{1}{2}\|\mathbf{A}_i\mathbf{y}_i-\mathbf{a}_i\|_2^2 \\
& \text{s.t.}
& & \mathbf{\underline{b}}_i \leq \mathbf{B}_i\mathbf{y}_i \leq \mathbf{\bar{b}}_i,
\end{aligned}
\end{equation}
where $\mathbf{A}_i$, $\mathbf{a}_i$, $\mathbf{B}_i$, $\mathbf{\underline{b}}_i$, $\mathbf{\bar{b}}_i$ define the equality and inequality constraints of the $i$-th task, respectively.
Then, we can obtain the solution of the $N_t$ tasks as
\begin{equation}
\mathbf{y}^* = \boldsymbol{\Lambda}(\mathbf{y}_n^*) = \mathbf{y}_1^* + \cdots + \prod_{k=1}^{i-1} \mathcal{N}(\mathbf{A}_k)\mathbf{y}_i^* + \cdots+ \prod_{k=1}^{N_t-1} \mathcal{N}(\mathbf{A}_k)\mathbf{y}_n^*,
\end{equation}
where $\mathcal{N}(\mathbf{A})$ is the null-space projector of the $\mathbf{A}$ matrix, and $\mathbf{y}_i^*$ is computed by solving the following quadratic program:
\begin{equation}\label{eq:hqp}
\begin{aligned}
& \underset{\mathbf{y}_i}{\min}
& & \frac{1}{2}\left\|\mathbf{A}_i\left[\boldsymbol{\Lambda}(\mathbf{y_{i-1}^*})-\prod_{k=1}^{i-1} \mathcal{N}(\mathbf{A}_k)y_i\right]-\mathbf{a}_i\right\|_2^2 \\
& \text{s.t.}
& &  \mathbf{\underline{b}}_i \leq \mathbf{B}_i\left[\boldsymbol{\Lambda}(\mathbf{y_{i-1}^*})-\prod_{k=1}^{i-1} \mathcal{N}(\mathbf{A}_k)y_i\right] \leq \mathbf{\bar{b}}_i, \\
&&& \hspace{4em}\vdots \\
&&& \mathbf{\underline{b}}_1 \leq \mathbf{B}_{1}\left[\boldsymbol{\Lambda}(\mathbf{y_{i-1}^*})-\prod_{k=1}^{i-1} \mathcal{N}(\mathbf{A}_k)y_i\right] \leq \mathbf{\bar{b}}_1.
\end{aligned}
\end{equation}
The above formulation does not include slack variables in the inequality constraints, although it is possible to use them as explained in~\cite{herzog-iros14}.
However, the use of quadratic costs to incorporate equality constraints is equivalent to including slack variables.

%% file: src/8_bios.tex
\begin{IEEEbiography}[{\includegraphics[width=1in,height=1.25in,clip,keepaspectratio]{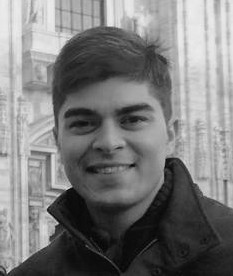}}]
    {Carlos Mastalli} received the M.Sc. degree in mechatronics engineering from the Simón Bolívar University, Caracas, Venezuela, in 2013 and the Ph.D. degree in bio-engineering and robotics from the Istituto Italiano di Tecnologia, Genoa, Italy, in 2017.
    
    He is currently a Research Associate in the University of Edinburgh and Alan Turing Institute, Edinburgh, U.K..
    From 2017 to 2019, he was a Postdoctoral Researcher in the Gepetto Team at LAAS-CNRS, Toulouse, France.
    Previously, he completed his Ph.D. on ``Planning and Execution of Dynamic Whole-Body Locomotion on Challenging Terrain'' under the supervision of I. Havoutis, C. Semini and D. G. Caldwell.
    He was Invited Researcher in ETH Zürich, Zurich, Switzerland, in 2016.
    His research combines the formalism of model-based approaches with the exploration of vast robot’s data for legged robotics.
    He has contributions in optimal control, motion planning, whole-body control and machine learning for legged locomotion.
\end{IEEEbiography}

\begin{IEEEbiography}[{\includegraphics[width=1in,height=1.25in,clip,keepaspectratio]{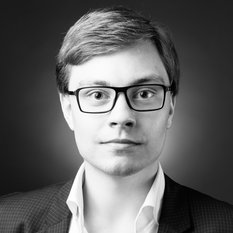}}]
    {Wolfgang Merkt} received the B.Eng.(Hns) degree in mechanical engineering with management and the M.Sc.(R) and Ph.D. degrees in robotics and autonomous systems from the University of Edinburgh, Edinburgh, U.K., in 2014, 2015 and 2019, respectively.
    
    He is currently a Postdoctoral Researcher at the Oxford Robotics Institute, University of Oxford.
    During his Ph.D., he worked on trajectory optimization and warm starting optimal control for high-dimensional systems and humanoid robots under the supervision of S. Vijayakumar.
    His research interests include fast optimization-based methods for planning and control, loco-manipulation, and legged robots.
\end{IEEEbiography}

\begin{IEEEbiography}[{\includegraphics[width=1in,height=1.25in,clip,keepaspectratio]{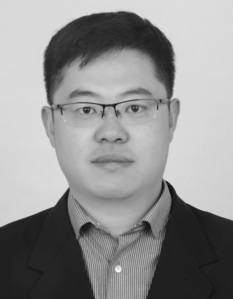}}]%
    {Guiyang Xin} received the M.Sc. degree in mechanical engineering from the China University of Geosciences, Wuhan, China, in 2012 and the Ph.D. degree in mechanical engineering from the Central South University, Changsha, China, in 2018.
    
    He is currently an Associate Professor at the Dalian University of Technology, Dalian, China.
    From 2018 to 2021, he was a Postdoctoral Researcher in the Institute of Perception, Action and Behaviour at the University of Edinburgh, Edinburgh, U.K..
    His research interests include legged robotics, impedance control, optimization-based control and teleoperation.
\end{IEEEbiography}

\begin{IEEEbiography}[{\includegraphics[width=1in,height=1.25in,clip,keepaspectratio]{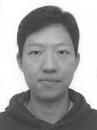}}]%
    {Jaehyun Shim} received the B.Eng. degree in mechanical engineering from the University of Tokyo, Tokyo, Japan in 2015 and the M.A.Sc. degree in mechanical engineering from the University of British Columbia, Vancouver, BC, Canada, in 2018.
    
    He is currently a Software Engineer at the University of Edinburgh.
    He worked on developing autonomous delivery robots as a Software Engineer at ROBOTIS, Seoul, South Korea.
    His research interests include optimization-based planning and legged robots.
\end{IEEEbiography}

\begin{IEEEbiography}[{\includegraphics[width=1in,height=1.25in,clip,keepaspectratio]{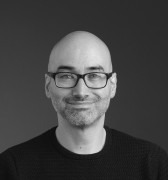}}]
    {Michael Mistry} received the B.S.E. in Mechanical Engineering \& Electrical Engineering from the University of Pennsylvania, USA, in 1998, the M.A. in Cognitive and Neural Systems from the Boston University, USA, in 2002 and the Ph.D. in Computer Science from the University of Southern California, USA, in 2009.

    He is a Professor and Personal Chair of Robotics at the School of Informatics, University of Edinburgh, and the Director of the Centre for Doctoral Training in Robotics and Autonomous Systems at the Edinburgh Centre for Robotics.
    His research focuses on the principles behind dexterous movement and control in robots and humans, particularly in environmental contact. Previously, Michael was a Senior Lecturer at the University of Birmingham, a postdoc at the Disney Research Lab at Carnegie Mellon University, a researcher at the ATR Computational Neuroscience Lab in Kyoto, Japan.
\end{IEEEbiography}

\begin{IEEEbiography}[{\includegraphics[width=1in,height=1.25in,clip,keepaspectratio]{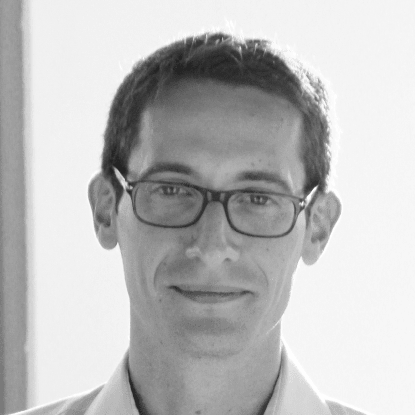}}]%
    {Ioannis Havoutis} received the M.Sc. in artificial intelligence and Ph.D. in Informatics degrees from the University of Edinburgh, Edinburgh, U.K.
    
    He is a Lecturer in Robotics at the University of Oxford.
    He is part of the Oxford Robotics Institute and a co-lead of the Dynamic Robot Systems group.
    His focus is on approaches for dynamic whole-body motion planning and control for legged robots in challenging domains.
    From 2015 to 2017, he was a postdoc at the Robot Learning and Interaction Group, at the Idiap Research Institute.
    Previously, from 2011 to 2015, he was a senior postdoc at the Dynamic Legged System lab the Istituto Italiano di Tecnologia.
    He holds a Ph.D. and M.Sc. from the University of Edinburgh.
\end{IEEEbiography}

\begin{IEEEbiography}[{\includegraphics[width=1in,height=1.25in,clip,keepaspectratio]{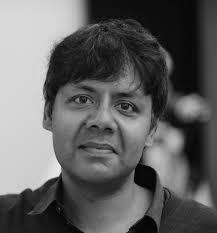}}]
    {Sethu Vijayakumar} received the Ph.D. degree in computer science and engineering from the Tokyo Institute of Technology, Tokyo, Japan, in 1998.
    
    He is Professor of Robotics and a Director with the Institute of Perception, Action, and Behaviour, School of Informatics, University of Edinburgh, U.K., where he holds the Royal Academy of Engineering Microsoft Research Chair in Learning Robotics.
    He also has additional appointments as an Adjunct Faculty with the University of Southern California, Los Angeles, CA, USA and a Visiting Research Scientist with the RIKEN Brain Science Institute, Tokyo.
    His research interests include statistical machine learning, robotics, planning and optimization in autonomous systems to human motor control, and optimality.
    Dr. Vijayakumar is a Fellow of the Royal Society of Edinburgh.
\end{IEEEbiography}